\documentclass{article}

\usepackage{PRIMEarxiv}

\usepackage[utf8]{inputenc} 
\usepackage[T1]{fontenc}    
\usepackage{url}            
\usepackage{booktabs}       
\usepackage{amsfonts}       
\usepackage{nicefrac}       
\usepackage{microtype}      
\usepackage{lipsum}
\usepackage{fancyhdr}       
\usepackage{graphicx}       

\usepackage[title]{appendix}
\usepackage{amsmath}
\usepackage{amsfonts}
\usepackage{amssymb}
\usepackage{graphicx}
\usepackage{siunitx}
\usepackage{xcolor}
\usepackage{booktabs}
\usepackage{caption}
\usepackage{float}
\usepackage{color,soul}
\usepackage{hyperref}
\hypersetup{
    colorlinks=true,
    linkcolor=blue,
    filecolor=blue,      
    urlcolor=blue,
    pdftitle={Overleaf Example},
    pdfpagemode=FullScreen,
    }

\usepackage{bm}

\usepackage{comment}

\title{Integrating Neural Operators with Diffusion Models Improves Spectral Representation in Turbulence Modeling}

\author{
  Vivek Oommen$^1$, Aniruddha Bora$^2$, Zhen Zhang$^2$, George Em Karniadakis$^{2}$\thanks{Corresponding author: george\_karniadakis@brown.edu}  \\
  1-School of Engineering \\
  2-Division of Applied Mathematics \\
  Brown University \\
  Providence, RI 02912\\
}

\begin{document}
\maketitle

\begin{abstract}
We integrate neural operators with diffusion models to address the spectral limitations of neural operators in surrogate modeling of turbulent flows. While neural operators offer computational efficiency, they exhibit deficiencies in capturing high-frequency flow dynamics, resulting in overly smooth approximations. To overcome this, we condition diffusion models on neural operators to enhance the resolution of turbulent structures. Our approach is validated for different neural operators on diverse datasets, including a high Reynolds number jet flow simulation and experimental Schlieren velocimetry. The proposed method significantly improves the alignment of predicted energy spectra with true distributions compared to neural operators alone. This enables the diffusion models to stabilize longer forecasts through diffusion-corrected autoregressive rollouts, as we demonstrate in this work. Additionally, proper orthogonal decomposition analysis demonstrates enhanced spectral fidelity in space-time. This work establishes a new paradigm for combining generative models with neural operators to advance surrogate modeling of turbulent systems, and it can be used in other scientific applications that involve microstructure and high-frequency content. See our project page: \url{https://vivekoommen.github.io/NO_DM/}
\end{abstract}

\keywords{Neural Operators \and  Diffusion Models \and Turbulence Modeling}

\section{Introduction}
\label{sec: intro}

Understanding turbulence is the last frontier in classical Physics. 
The complexity arises due to the high Reynolds numbers encountered in atmospheric and engineering turbulence, which induce an energy cascade producing spatiotemporal scales spanning multiple orders of magnitude, rendering them difficult to resolve fully.  Also, extreme sensitivity to important other parameters, such as body roughness and external flow perturbations have a compound effect. In 1972, Howard Emmons, a distinguished researcher from Harvard, famously asserted that turbulence would never be computable. However, with the advent of supercomputers and developing dealiased spectral methods, a breakthrough occurred just a couple of years later. Steven Orszag of MIT, utilizing the NCAR's first supercomputer (CDC7600), conducted the first direct numerical simulation (DNS) of homogeneous turbulence, albeit at lower Reynolds numbers (Re).

Progress has been made since then on simulating accurately several fundamental turbulent flows \cite{kim1987turbulence, choi1994active, du2000suppressing} but even today with exascale supercomputers and a speed-up of over $10^{10}$ over the CDC7600, DNS of turbulence is still limited to relatively low Reynolds numbers and simple-geometry flows.
On the experimental side, methods such as particle image velocimetry (PIV) and particle tracking velocimetry (PTV) can provide useful information but cannot lead to quantitative understanding as DNS does, e.g. compute all critical statistical quantities of interest, and they may have difficulty resolving high wavenumbers and high frequencies.

It is clear, therefore, that a fundamentally different approach is needed to break the deadlock. What is new now that was not available before is the  {\em in-situ} diagnostic tools, such as PIV and PTV, that can provide at least some partial information about the flow dynamics both in time and space. However, existing simulation methods are based on conventional numerical solvers, e.g. spectral methods, that require elaborate data assimilation techniques to integrate measurements. To this end, the recent emergence of physics-informed neural networks (PINNs)  \cite{raissi2019physics, karniadakis2021physics, toscano2024pinns} enable seamless integration of multimodal data, and offer new opportunities for drastically advancing turbulence modeling. Indeed, PINNs have been very effective in discovering new physics even from sparse data by employing the so-called ``gray-box" models. In this new paradigm, we postulate a mathematical model, e.g., the Navier-Stokes equations (NSE) but with an unknown correction term represented by a neural network (NN), whose weights and biases can be computed using the available data. Subsequently, the discovered NN can be expressed analytically using modern symbolic regression techniques, e.g. see \cite{kiyani2023framework,zhang2024discovering,ahmadi2024ai, de2023ai,cavity2024physics}, and new parameterization using fractional calculus to account for non-local interactions  \cite{Fangying-GK}. 


Another approach in assimilating data and
provide faster solutions than DNS, which is prohibitively expensive at high Reynolds numbers, is through neural operators.
%
%
In particular, classical DNS provides the solution for a given set of conditions. If there is a change in one of the given conditions, the solvers have to re-run which further adds to the computational cost. To circumvent this issue, neural operators were developed to work with a plurality of operating conditions and even different parameters, e.g. the Reynolds number $Re$~\cite{lu2021learning,li2020fourier,raonic2023convolutional,li2022transformer,ovadia2023ditto,sharma2024graph}. In a neural operator the objective is to learn the mapping between two or more infinite-dimensional functional spaces. Although training a neural operator offline is expensive, the trained neural operator can infer solutions of unseen conditions almost instantaneously online, with negligible computational costs. Hence, neural operators can serve as efficient surrogates for different applications \cite{oommen2024rethinking,bora2023learning,oommen2022learning,peyvan2024riemannonets,hao2023gnot,li2022transformer,zappala2024learning,ye2024locality, wan2025deepvivonet}.

Despite the promising results, inference of nonlinear dynamics and its temporal evolution using neural operators suffer from over-smoothing and loss of its high frequency content, a critical property of turbulent flows.  
This can also be attributed to the spectral bias of neural networks \cite{xu2025understanding}, which is the behavior of neural networks to learn only the low frequencies while remaining oblivious to higher frequency components. 
The spectral bias of neural operators/networks becomes increasingly significant in the surrogate modeling of turbulent systems whose energy spectrum decays rather slowly leading to non-trivial energy levels at higher wave numbers. 
Some of the recent efforts related to mitigating spectral bias suffered by neural networks include multi-scale DNN \cite{cai2019multi}, phase-shift DNN \cite{cai2020phase} and the use of Fourier feature mappings \cite{tancik2020fourier, wang2021eigenvector}.
In the context of neural operators, PDE-Refiner \cite{lippe2024pde} relies on iterative refinements, \cite{zhang2022hybrid} blends neural operators with relaxation methods and \cite{chakraborty2025binned} proposes a novel binned spectral power loss-based training to address the spectral limitations.

In the current work, we investigate the use of generative AI models to overcome the spectral bias barrier.
Surveying the recent advances in generative modeling demonstrates that generative adversarial networks (GANs) \cite{goodfellow2014generative}, normalizing flows \cite{rezende2015variational} or diffusion models \cite{sohl2015deep,ho2020denoising} are capable of generating high-quality images with spectral properties that are desirable in turbulence modeling. 
Specifically, herein we aim to utilize the diffusion models to mitigate the spectral bias in neural operators. 
Diffusion models use gradual denoising of random noise to generate samples from the target distribution. 
Since diffusion models reverse the corruption process one step at a time, during the generative process they predict  the next higher frequency based on the previously generated lower-frequency component. 
This is analogous to auto-regression in frequency space, and the generated images demonstrate better alignment across the frequency spectrum. 

Recently, diffusion models have seen rapid development. 
Ho \emph{et al.} \cite{ho2020denoising} introduced the Denoising Diffusion Probabilistic Models (DDPM), which involves corruption of the signal with Gaussian noise followed by sequential denoising for retrieving signal at least from within the same distribution.
Although DDPMs achieve high sample quality, the computational cost is high due to its requirement of a large number of sampling steps, typically 1000. 
Song \emph{et al.} \cite{song2020score} demonstrated improved training and sampling by using neural networks to accurately estimate the score, which is a denoising-time-dependent gradient field of the perturbed data distribution, and then using numerical stochastic differential equation solvers such as Langevin MCMC (Markov Chain Monte Carlo) to generate samples.
Karras et al. \cite{karras2022elucidating}, performed a systematic investigation that helped elucidate the training and design of diffusion models by analyzing all types of diffusion models from within a common framework. 
The noise-weighted loss during training and improved sampling routines using preconditioned networks helped to bring down the number of sampling steps from $\approx 1000$ in DDPM to $\approx 30$.
 Furthermore, \cite{saharia2022palette} demonstrated the efficiency of diffusion models on a variety of image-to-image tasks. 
 For example, in the image colorization problem, a grayscale image serves as a prior for the diffusion model that generates the colorized version. 
 Moreover, in the image inpainting challenge, the masked version of the image serves as an effective prior to the diffusion model that generates the full image. 
 Similarly, \cite{robertson2023local, robertson2024micro2d, buzzy2024statistically} used statistically conditioned diffusion models for generating heterogeneous and statistically diverse microstructures.
 \cite{dong2024data} builds stochastic and non-local closure models using conditional diffusion models and neural operators. 
 \cite{huang2024diffusionpde, kohl2024benchmarking, wang2022deep, molinaro2024generative, lockwood2024generative} also demonstrate applications of generative AI in scientific modeling. 

The ability of diffusion models to generate images with a slowly decaying energy spectrum, and the efficient utilization of priors motivated us to investigate the effectiveness of training a diffusion model conditioned on the neural operator's output as their prior. 
First, we demonstrate that the neural operators suffer from spectral bias by illustrating predictions with over-smoothening. 
We mitigate the over-smoothening issue using the score-based diffusion model conditioned on the neural operator's output. 
The improved spectral properties can be directly observed in our visualizations and the energy spectra plots, where we compare the ground truth and the neural operator's predictions with the output of the conditioned diffusion models. 
Our approach is shown to work across a variety of neural operator architectures.
We demonstrate the effectiveness of our framework on test cases with Reynolds numbers ranging from 2000 to $10^6$, 2D \& 3D large-eddy simulations (LES), and raw experimental Schlieren velocimetry datasets.   
Furthermore, we perform a proper orthogonal decomposition (POD) analysis to provide a quantitative comparison of the energy decay and the POD modes of the predictions with the ground truth. Our ultimate objective is to accelerate DNS and LES by interweaving bursts of DNS/LES with the new combined method as well as to infer turbulent flow evolution at as DNS-level fidelity based on sparse experimental data.

\section{Results}
\label{sec: results}

In the context of modeling physical systems, a non-linear mathematical operator $\bm{\mathcal{N}}$ that governs the evolution of a spatio-temporal field $\bm{u}(\bm{x},t)$, can be expressed as,

\begin{equation}
    \bm{u}(\bm{x}, t+\Delta t) = \bm{\mathcal{N}}(\bm{u}(\bm{x}, t)) 
\end{equation}

Unlike the conventional discretization-based solvers, neural operators ($\bm{\mathcal{G}}$) that learn to approximate the underlying true operator ($\bm{\mathcal{N}} \approx \bm{\mathcal{G}}$) from its solutions, have been shown to generalize well across the unseen initial conditions.

\begin{figure}
  \centering
  \includegraphics[width=0.99\textwidth]{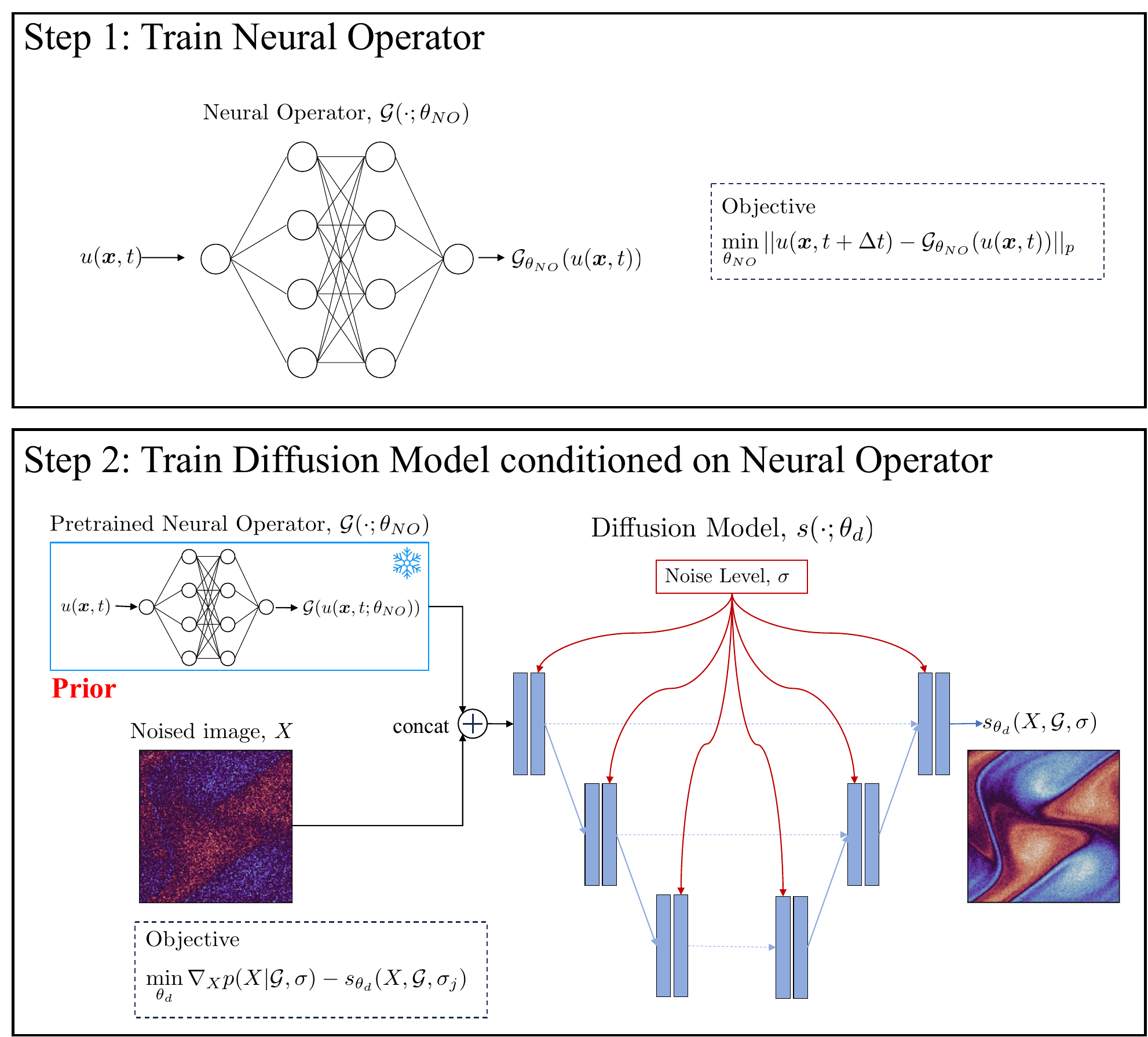}
  \caption{\textbf{Methodology.} First, we train a neural operator to learn the mapping between the function spaces by minimizing, typically, the $L^2$ norm with respect to available ground truth. 
  Next, we train a diffusion model conditioned on a neural operator's output to approximate the ground truth data distribution using denoising score matching. }
  \label{fig: methodology}
\end{figure}

However, the neural operators often learn only the low wave numbers and ignore the higher wave numbers leading to an overall smoother approximation of the true solution. 
The issue is rooted in the spectral bias associated with the conventional training of the neural networks.
In this section, we present the results of the experiments we performed to investigate whether diffusion models can help mitigate the issue of spectral bias in neural operators.
The detailed methodology is explained in \autoref{sec: Methods} and is illustrated in \autoref{fig: methodology}.
All the neural operators and diffusion models used in this study have approximately 2 to 3 million trainable parameters.

\subsection{Kolmogorov Flow}


The unsteady two-dimensional incompressible Navier-Stokes equation for a viscous, incompressible fluid with Kolmogorov forcing \cite{chandler2013invariant} in the vorticity form is given by:
\begin{equation}\label{eq:NS_equation}
\begin{cases}
    \partial_t \omega + \bm{u} \cdot \nabla  \omega = \nu \Delta \omega + f(x,y), & (x,y) \in (0,2\pi)^2, t \in (0, t_{final}] \\
    f(x,y) = \chi ( \sin(2\pi(x+y)) + \cos(2\pi (x+y)) ), & (x,y) \in (0,2\pi)^2 \\
    \nabla \cdot \bm{u} = 0, & (x,y) \in (0,2\pi)^2, t \in (0, t_{final}] \\
    \omega(x,y,0) = \omega_0, & (x,y) \in (0,2\pi)^2 
\end{cases}
\end{equation}
where $\chi\!=\!0.1$, $\omega$ is the vorticity, $\bm{u}$ is the velocity field vector,  $\nu(=10^{-5})$ is the kinematic viscosity, and $\Delta$ is the two-dimensional Laplacian operator. 
We consider periodic boundary conditions. The source term $f$ is given by $f(x,y) = 0.1 (\sin(2\pi (x + y)) + \cos(2\pi(x + y)))$, and the initial condition $\omega_0(x)$ is sampled from a Gaussian random field according to the distribution $\mathcal{N}(0, 14^{1/2} (-\Delta + 196 I)^{-1.5})$.
We used the publicly available pseudo-spectral solver \cite{li2020fourier} to generate a dataset with 1000 samples, and randomly split it to train, validation and test sets in the ratio 80:10:10. 
The dataset originally generated with a spatial resolution of $512 \times 512$, was downsampled to $128 \times 128$ to train the neural operators and diffusion models. 

\begin{figure}[!h]
  \centering
  \includegraphics[width=0.99\textwidth]{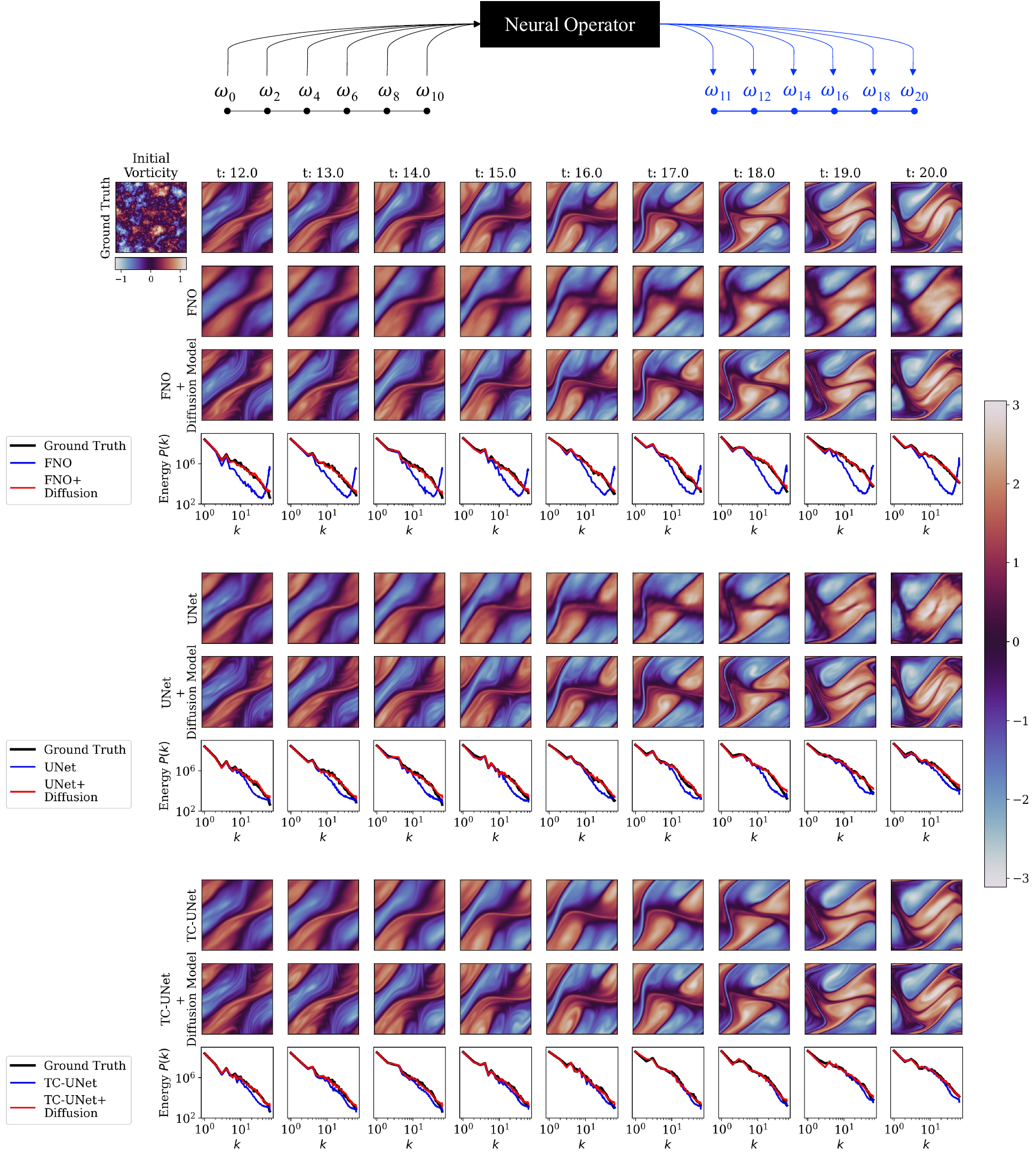}
  \caption{\textbf{Kolmogorov Flow.} The first row shows the vorticity field in a kolmogorov flow simulated using a pseudo-spectral solver.
  The following rows show the predictions of different types of neural operators - FNO, UNet \& TC-UNet and that of a score-based diffusion model conditioned on the respective neural operator as a prior.
  The corresponding energy spectra at each timestep are also displayed.
  }
  \label{fig: kolmogorov}\vspace{-0.2in}
\end{figure}

We train the neural operator ($\mathcal{G}$), to learn the mapping $\omega(x,y,t)_{|t\in[0,10]} \rightarrow \omega(x,y,t)_{|t\in[10, t_{final}]}$ of the vorticity field upto $t=10$ to $t_{final}=20$. 
We borrow the same problem setup as in \cite{li2020fourier}, but with less smooth initial conditions (see the initial vorticity in \autoref{fig: kolmogorov}), causing higher spatial gradients in the resulting simulation as demonstrated in the first row of \autoref{fig: kolmogorov}. 
We investigate the effectiveness of the proposed framework across different neural operator architectures - Fourier Neural Operator (FNO) \cite{li2020fourier}, UNet, and Time-Conditioned UNet (TC-UNet) \cite{ovadia2023ditto}.
The FNO accurately learns the dominant low wave-number modes, but remains oblivious to the higher wavenumbers corresponding to the finer spatial features. 
Additionally, the presence of high spatial gradients in the true solution causes the FNO predictions to exhibit Gibbs oscillations.
The UNet and TC-UNet also show smoothening in their predictions. 

The spectral characteristics of the images generated by diffusion models \cite{croitoru2023diffusion, yang2023diffusion} motivated us to utilize them to improve neural operator predictions.
Specifically, we train a score-based diffusion model \cite{karras2022elucidating} conditioned on the neural operator's output, to approximate the ground truth data distribution using denoising score-matching.
Once trained, the diffusion model uses annealed Langevin dynamics-based sampling to generate its prediction conditioned on the specific neural operator output.  
\autoref{fig: kolmogorov} illustrates that the vorticity field estimated by the conditioned diffusion model gets rid of the Gibbs oscillations (for FNO), recovers the finer features, and agrees well with the true power spectrum for all the neural operators compared here. 
Furthermore, we perform POD to analyze the energy decay spectrum and to visualize \& compare the POD modes of the Ground Truth, FNO, and FNO + Diffusion Model in \autoref{fig: pod_kolmogorov}.
The proposed framework demonstrates a better alignment with true modes than the neural operator alone.

\begin{figure}[!h]
  \centering
  \includegraphics[width=0.85\textwidth]{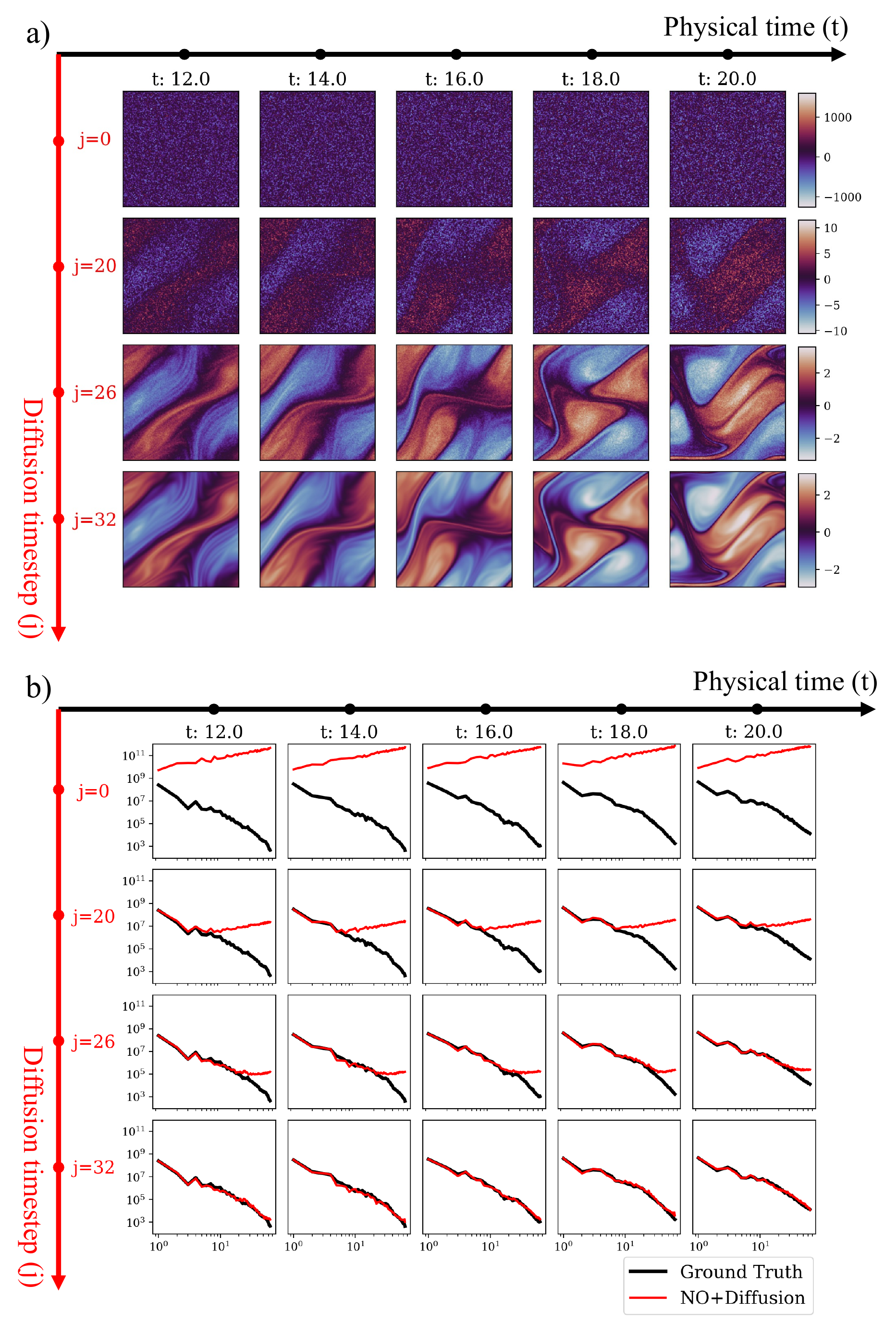}
  \caption{\textbf{Sampling Process.} a) An illustration of the sampling process in a score-based diffusion model conditioned on the Fourier neural operator's prediction (NO + Diffusion), and b) the comparison of the corresponding energy spectra with the ground truth. 
  }
  \label{fig: diffusion_process}\vspace{-0.2in}
\end{figure}

\subsection{Buoyancy-driven Transport}

Here we train a neural operator to learn the buoyancy-driven transport of a scalar concentration field ($d$) \cite{thuerey2021physics}. 
We can define the buoyancy model with Boussinesq approximation using the incompressible Navier-Stokes equation accompanied by the advection transport of $d$ and can be expressed in terms of the buoyancy factor $\xi$ as,

\begin{align}
    \frac{\partial \bm{u}}{\partial t} + \bm{u} \cdot \nabla \bm{u} &= -\frac{1}{\rho} \nabla P + \nu \nabla^2 \bm{u} + [0,1]^T \xi d \quad \text{ s.t } \quad \nabla \cdot \bm{u}=0 \\
    \frac{\partial d}{\partial t} + \bm{u} \cdot \nabla d &= 0
\end{align}
where $0 \leq t \leq 21$, $\nu=10^{-2}$ and $\xi=0.5$. 
The system has Dirichlet boundary conditions for the velocity ($\bm{v}=0$) and Neumann boundary conditions for the concentration field ($\frac{\partial d}{\partial x} = 0$).
We utilized the dataset from PDEArena \cite{gupta2022towards} comprising of 2080 
train, 260 validation and 260 test trajectories, generated using $\Phi$flow \cite{holl2024phiflow}. 
Each simulation was saved every $\Delta t = 1.5$, resulting in 14 time steps per trajectory.

We train a U-Net as the neural operator to learn the evolution of $d$. The neural operator takes the states at 4 previous timesteps as the input and predicts the next time step as output. 
During training, the ground truth data is provided as the input. 
But while validating and testing, the ground truth data at only $t= 0, \Delta t, 2 \Delta t, 3 \Delta t$ is available. 
So the neural operator performs an auto-regressive rollout to infer the future states. 
The setup of the neural operator is consistent with that demonstrated in \cite{gupta2022towards}. 

\begin{figure}[!h]
  \centering
  \includegraphics[width=0.99\textwidth]{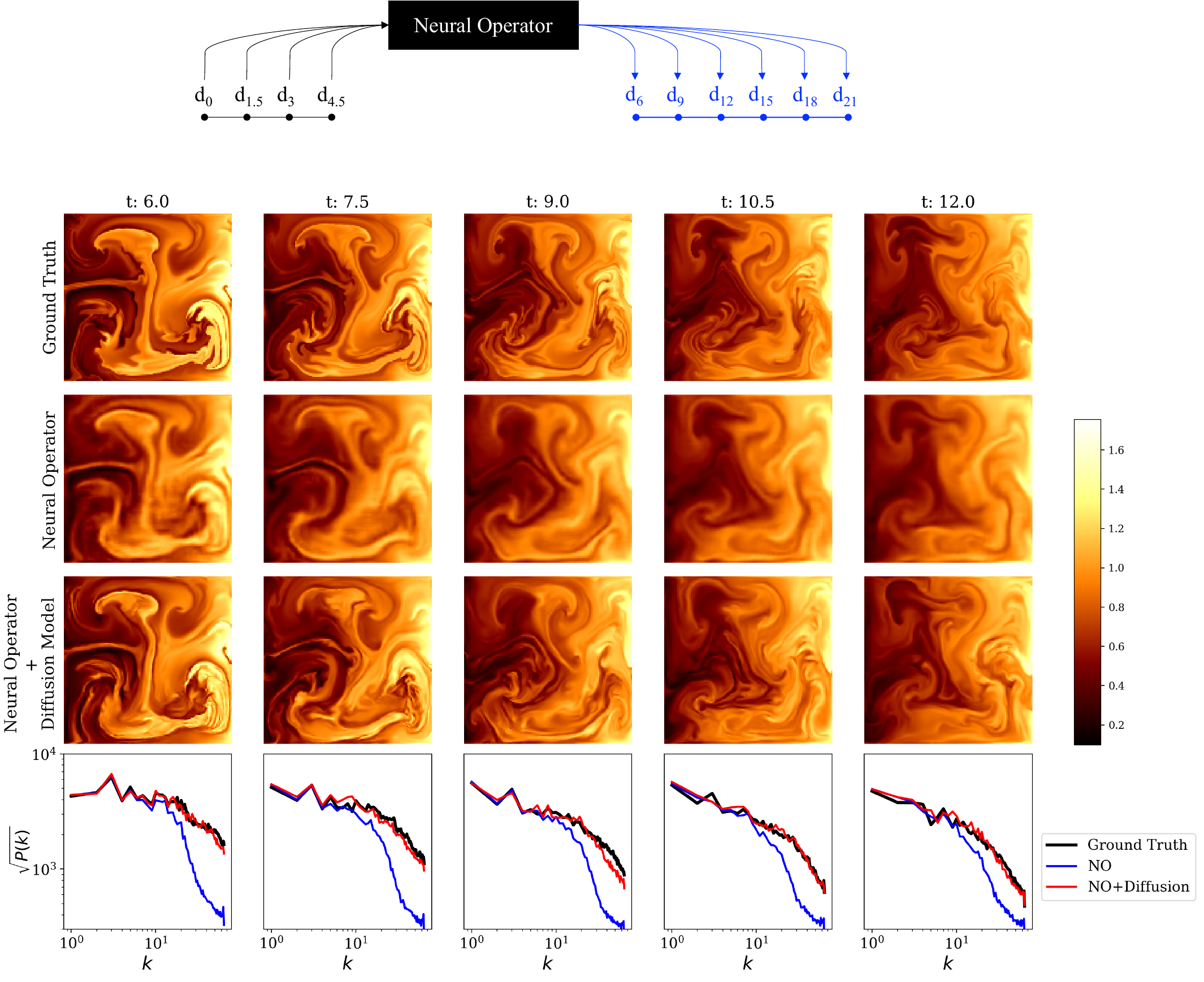}
  \caption{\textbf{Buoyancy-driven Transport.} The first row illustrates the true transport of the concentration field $d$ simulated using $\Phi$Flow. 
  The second row is the prediction of the (UNet) neural operator.
  The third row visualizes the prediction of the diffusion model conditioned on the neural operator's output.
  The last row compares the square roots of the respective energy spectra to highlight the differences. 
  }
  \label{fig: buoyancy}
\end{figure}

In \autoref{fig: buoyancy}, we visualize the ground truth simulation in the first row, predictions of the neural operator in the second row, and that of the diffusion models conditioned on the neural operator's output in the third row.  
We compare the spectral distribution of each of the simulations in row 4. 
We can observe the over-smoothening in the fields predicted by the neural operator. 
The diffusion model conditioned on the neural operator achieves better spectral similarity with the ground truth, compared to the neural operator alone. 
Additionally, we investigate the impact of the neural operator on the diffusion model's prediction by varying the number of trainable parameters of the neural operator in \autoref{sec:discussion}.
Furthermore, we perform POD to analyze the energy decay spectrum and to visualize \& compare the POD modes of the Ground Truth, UNet, and UNet + Diffusion Model in \autoref{fig: pod_buoyancy}.
The proposed framework demonstrates a better alignment with true modes than the neural operator alone. 
Although the buoyancy-driven flow with $\nu=10^{-2}$ is not an example of a turbulent system, this test case effectively demonstrates the spectral bias suffered by neural operators/networks towards low frequencies.

\subsection{Turbulent airfoil wake}

Here we consider a turbulent wake flow downstream of a NACA0012 airfoil with a chord length of $L_c$ at a Reynolds number of 23000, a free stream Mach number of 0.3 and an angle of attack of $6^\circ$ \cite{yeh2019resolvent}.
The flow displays coherent structures linked to the Kelvin-Helmholtz instability over the separation bubble and von Karman vortex shedding in the wake, while preserving the broad complexity typical of turbulent flow \cite{towne2023database}. 
This makes it an excellent platform for CFD validation, flow analysis, and the investigation of reduced-complexity models.
We directly utilize the large eddy simulation (LES) dataset generated by Towne \emph{et al.} \cite{towne2023database} from "Deep Blue Data", a publicly available database from the University of Michigan. 
The three-dimensional flow was simulated using a finite-volume based compressible solver - \textit{CharLES} \cite{bres2017unstructured}.
The LES dataset is comprised of 16,000 time-resolved snapshots collected at a constant convective time increment of $\tau = 0.0104\frac{L_c}{U_{\infty}}$. For more details on the problem setup and dataset generation, please refer to section VII in \cite{towne2023database}.

\begin{figure}[!h]
  \centering
  \includegraphics[width=0.99\textwidth]{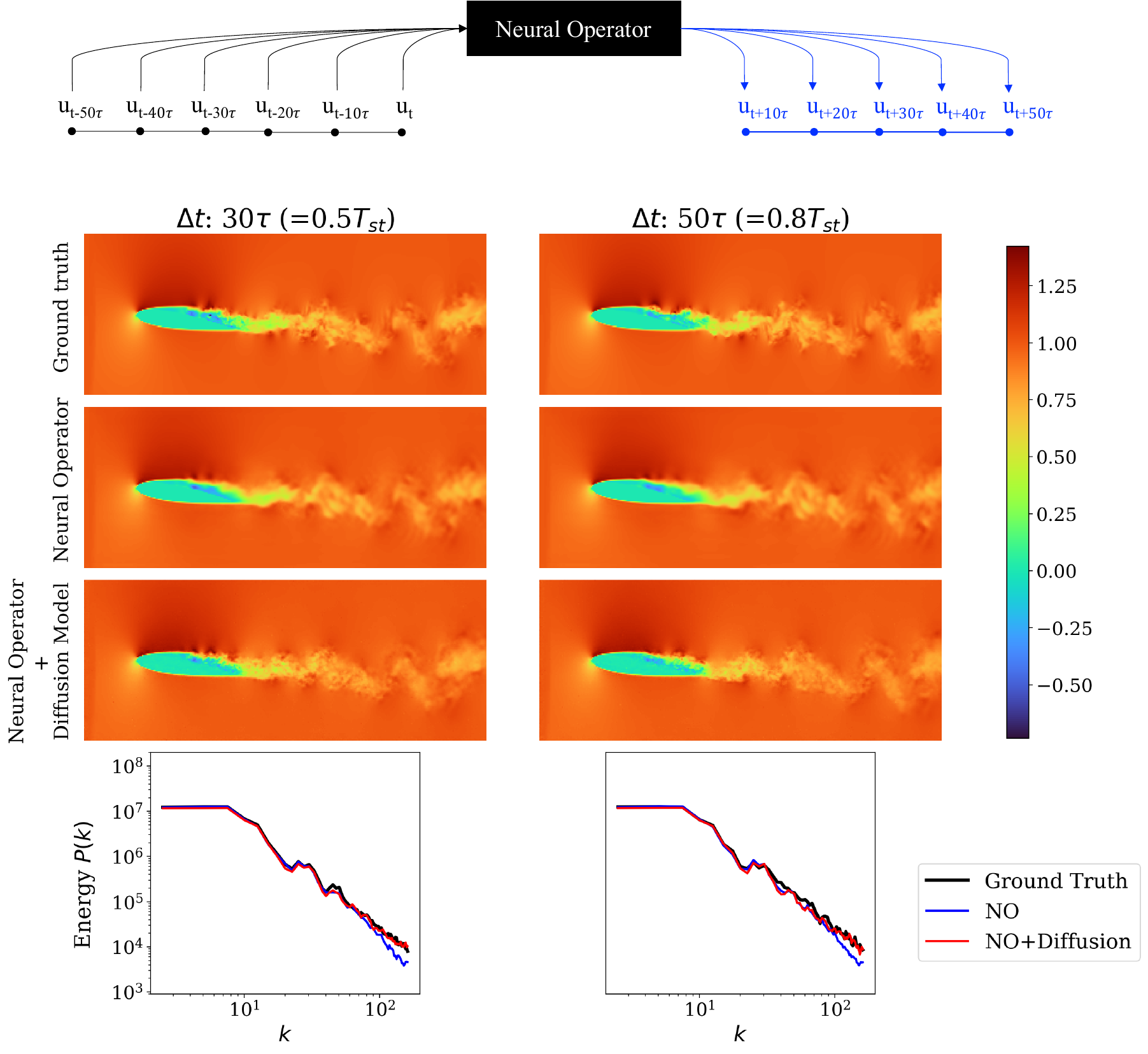}
  \caption{\textbf{Turbulent wake of NACA0012.} Distribution of the streamwise velocity $u$ at the mid-span slice at $Re=23,000$.
  The first row is the ground truth generated from LES, second row is the prediction of neural operator (TC-UNet) and the third row is the prediction of a diffusion model conditioned on the neural operator's output. 
  The last row plots the energy spectrum of the ground truth and predictions. }
  \label{fig: airfoil}
\end{figure}

For training the neural operator and the diffusion model we consider  only a subset of the streamwise velocity component ($u$) along the mid-span airfoil slice of the aforementioned 3D simulation.
We consider 1000 timesteps, at a time increment of $5\tau$, within a subdomain $4L_c \times 2.5L_c$.
We train a UNet-based neural operator, TC-UNet \cite{ovadia2023real}, that takes a history of states - $\{ u_{t-50\tau}, u_{t-45\tau}, ..., u_t \}$  -  as the input and directly predicts the states - $\{ u_{t+5\tau},  u_{t+10\tau}\}$.
During training, the ground truth from the LES is provided as the input to the neural operator. 
However, during validation and testing, we carry out an autoregressive rollout by feeding back the states predicted by the neural operator as the input to predict all the states till $t+50\tau$ that corresponds to 80\% of the Strouhal time period ($T_{st}$). 
We split the 1000 timesteps into train, validation and test datasets in the ratio 80:10:10.
From \autoref{fig: airfoil}, we observe that the neural operator predicts a smoothened representation of the true LES simulations. 
The diffusion model conditioned on the neural operator's output can recover the finer features, with a better aligned power spectrum.
Furthermore, we perform POD to analyze the energy decay spectrum and to visualize \& compare the POD modes of the Ground Truth, TC-UNet, and TC-UNet + Diffusion Model in \autoref{fig: pod_airfoil}.
The diffusion model conditioned on the neural operator demonstrates a better alignment with true modes than the neural operator alone.

\subsection{3D Turbulent Jet}

A jet is a free shear flow consisting of a fast inner stream and a slower outer stream \cite{towne2023database}. 
In a jet, a high-velocity stream of fluid is ejected through a nozzle into the surrounding medium where the mean velocity gradient is not affected by the walls.
Simulating the jet facilitates understanding the flow dynamics and optimizing designs in real-world applications in aerospace, automotive and energy industries. 
Here we try to model a subsonic ($M=0.9$), isothermal, turbulent ($Re \approx 10^6$) jet from a convergent-straight nozzle using neural operators. 

Again we directly utilize the dataset generated by Towne \emph{et al.}, which is publicly available through the Deep Blue Data repository from the University of Michigan. 
The simulations were performed with a high-fidelity LES-based compressible flow solver \textit{CharLES} \cite{bres2017unstructured} using the Finite Volume Method. 
For more details about the problem setup and configurations, please refer section II in \cite{towne2023database}. 
The original dataset consists of 10000 snapshots of the jet sampled every 0.2 acoustic time units $\tau = 0.2 t c_{\infty}/D$, where $c_{\infty}$ is the free stream speed of sound and $D$ is the nozzle exit diameter. 

For training the neural operator we consider the first 1000 timesteps of the velocity field vector ($\bm{u}$) comprising of the axial ($u_x$), radial ($u_r$) and azimuthal ($u_{\theta}$) components, and utilize the first 800 for training, the next 100 for validation, and the next 100 for testing. 
The dataset domain is $11.71 D$ along the axial direction and $2.78 D$ along the radial direction. 
The neural operator, TC-UNet, is trained to learn the mapping $\bm{u}(\bm{x},t) \rightarrow \bm{u}(\bm{x}, t + \Delta t)|_{\Delta t \in [0,10\tau]}$.

\begin{figure}
  \centering
  \includegraphics[width=0.85\textwidth]{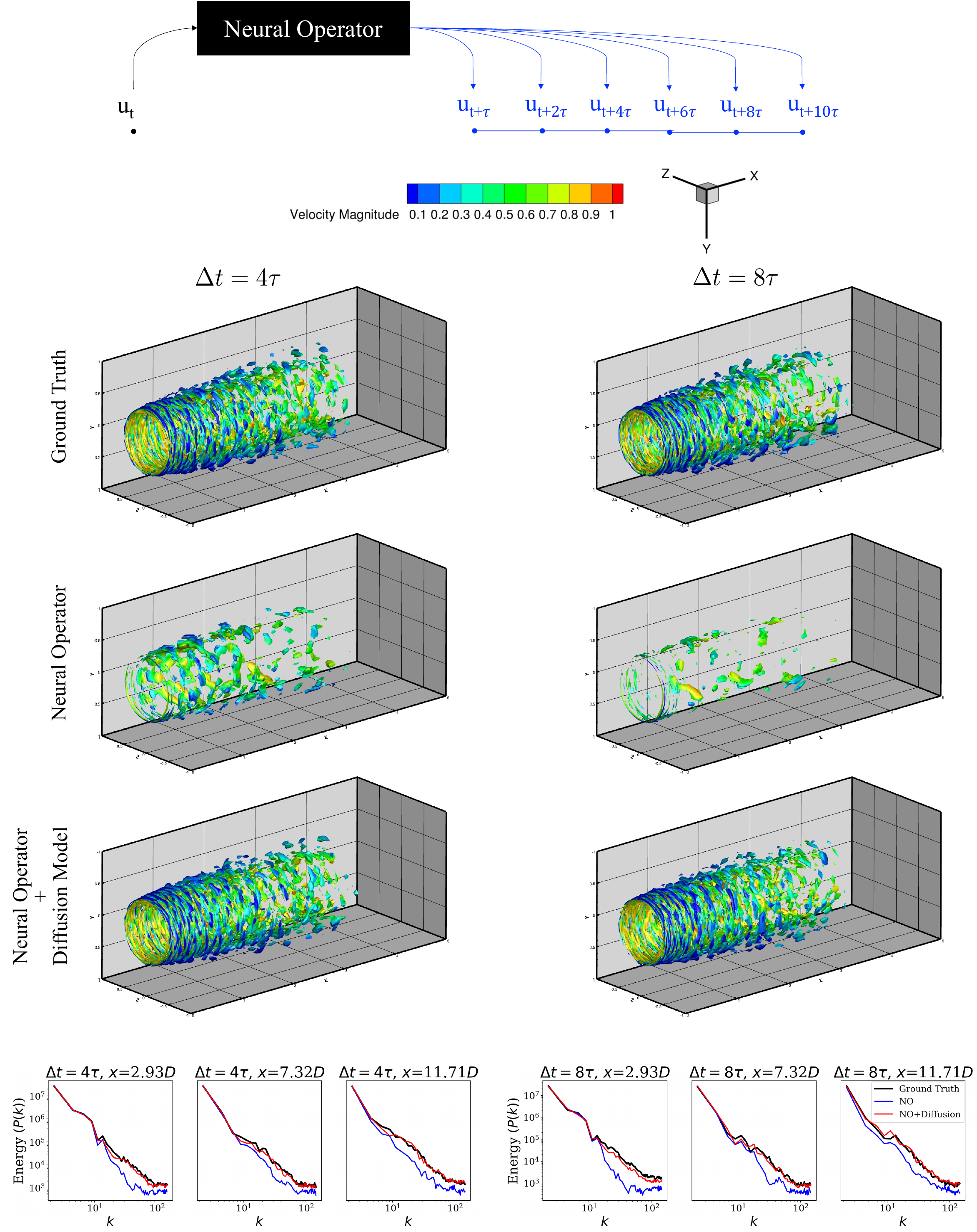}
  \caption{\textbf{3D turbulent jet.} Isosurfaces corresponding to Q-criterion = 1.5 of a turbulent jet ($M=0.9$, $Re=10^6$) is visualized. The color of the isosurfaces corresponds to the magnitude of the velocity. The first three rows represent solutions of the reference LES, neural operator (TC-UNet), and diffusion model conditioned on the neural operator. The last row compares the energy spectra at different axial cross-sections corresponding to $\Delta t = 4\tau, 8\tau$.
  }
  \label{fig: jet_3d}
\end{figure}

We illustrate the isosurfaces corresponding to Q-criterion=1.5  in \autoref{fig: jet_3d}. 
The neural operator is unable to retain the higher wavenumbers in its predictions, especially at the later time states, but captures the general trend of the system contained in the lower wavenumbers.
This facilitates the neural operator to behave as an effective prior to the diffusion model which recovers the high frequency components, from angular cross-sections of the velocity vector field inferred by the neural operator. 
This phenomenon is also illustrated using the power-spectrum plots of the magnitude of the velocity field at different axial cross-sections.
Additionally, we perform POD to analyze the energy decay spectrum and to visualize \& compare the POD modes of the Ground Truth, TC-UNet, and TC-UNet + Diffusion Model in \autoref{fig: pod_jet_3d}.
The diffusion model conditioned on the TC-UNet demonstrates a better alignment with true modes than the TC-UNet alone. 

\subsection{Schlieren velocimetry of turbulent helium jet in air}

So far our results demonstrate the effectiveness of using neural operators as priors of diffusion models, on numerically simulated turbulent databases. 
In the last test case, we investigate if the proposed framework can be directly applied to real experimental setups.

We train the surrogate network to model a turbulent helium jet in air from Schlieren velocimetry data \cite{settles2022schlieren}.  

\begin{figure}
  \centering
  \includegraphics[width=0.8\textwidth]{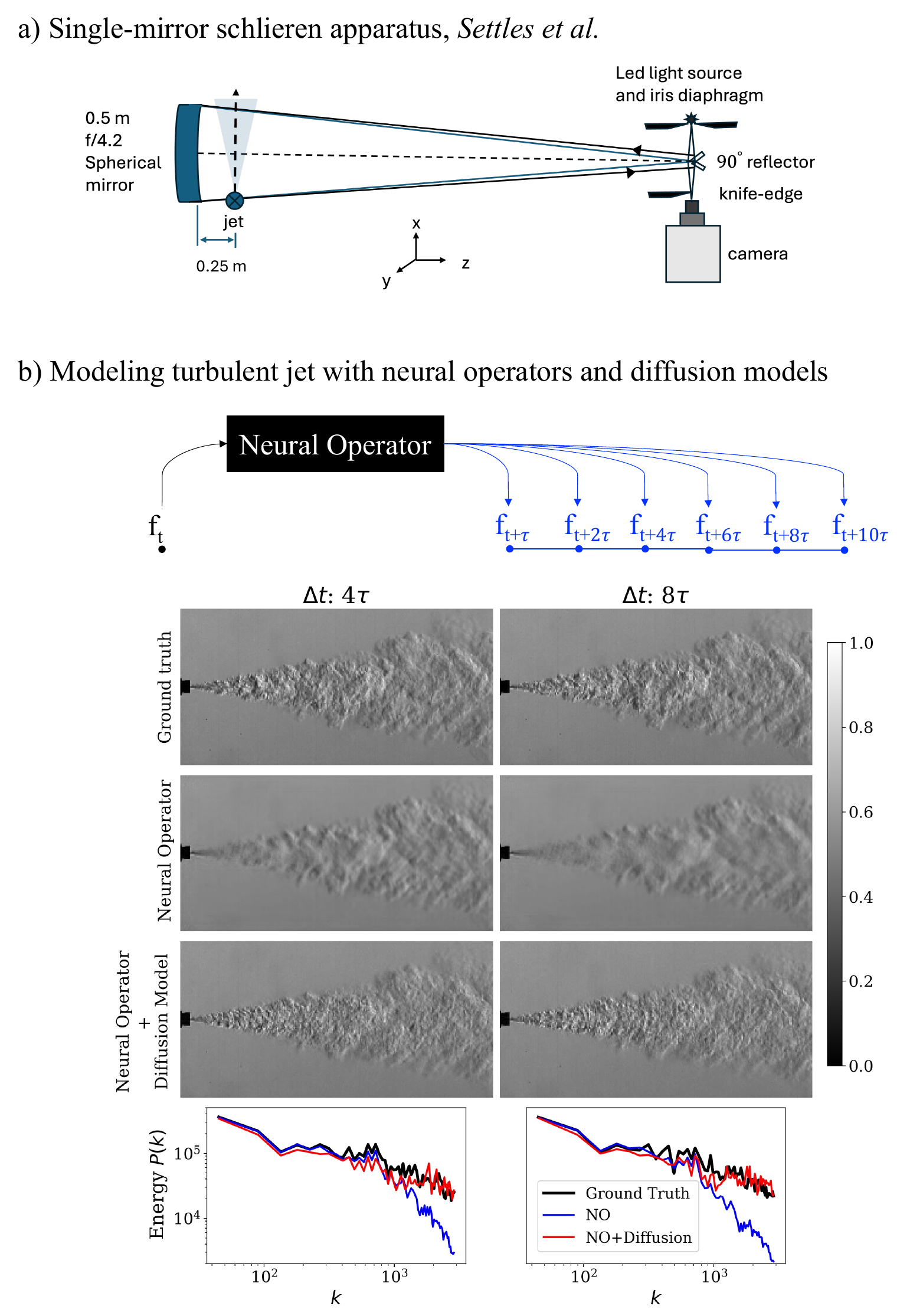}
  \caption{\textbf{Schlieren velocimetry. } a) Schematic of knife-based single mirror Schlieren apparatus (screenshot from \cite{settles2022schlieren}) Settles \emph{et al.} uses to visualize turbulent jet of helium in the air, with a jet exit $M=0.45$ and $Re=5890$. 
  b) The density gradients at $\Delta t = 4\tau, 8\tau$ captured using the schlieren velocimetry apparatus (row 1), estimated by neural operator, TC-UNet (row 2) and estimated using diffusion model conditioned on neural operator in (row 3). 
  In the last row, we compare the energy spectra of the respective snapshots captured/estimated by the three approaches. }
  \label{fig: schlieren}
\end{figure}

The system under consideration is a round turbulent helium jet in air expelled from a nozzle of diameter $d$ at a jet exit mach number of $M=0.45$, velocity of $u=436 m/s$ and Reynolds number $Re = 5890$. 
The domain of interest is $200d$ along the x-direction and $100d$ along the y-direction. 
A schematic of the traditional single mirror schlieren apparatus is provided in \autoref{fig: schlieren}a (screenshot from \cite{settles2022schlieren}).
A Photron APX-RS camera captures the images of the turbulent jet at 6000 frames/s with an image exposure of 167 $\mu s$.

We directly utilize the publically available schlieren velocimetry dataset provided by \cite{settles2022schlieren}, in raw .tif format. 
Each snapshot is separated by $\tau = 1/6000 s$.
For training the neural operator we consider 1000 snapshots of the density gradient field ($f$), and utilize the first 800 for training, next 100 for validation and the last 100 for testing. 
The neural operator, TC-UNet, is trained to learn the mapping $f(\bm{x},t) \rightarrow f(\bm{x}, t + \Delta t)|_{\Delta t \in [0,10\tau]}$, and the diffusion model takes the output of TC-UNet as its prior. 

\autoref{fig: schlieren} a) illustrates the experimental setup of a traditional single-mirror schlieren apparatus used in \cite{settles2022schlieren}. 
In part b) the first row represents the density gradient captured using the schlieren velocimetry setup at $\Delta t  = 4\tau, 8\tau$, with nozzle on the left and helium expelled towards right along the x-axis. 
The second and the third row represents the predictions of neural operator and the diffusion model conditioned on neural operator, again at the same time steps. 
The color bar represents the density gradient normalized between 0 and 1. 
In the last row we compare the energy spectra of as a function of wavenumber for the three approaches discussed for both the timesteps. 
From both the contour plots as well as the energy spectrum, the predictions of the diffusion model conditioned on the neural operator align better with the ground truth than the neural operator alone.
We also perform POD to analyze the energy decay spectrum and to visualize \& compare the POD modes of the Ground Truth, TC-UNet, and TC-UNet + Diffusion Model in \autoref{fig: pod_schlieren}.
The diffusion model conditioned on the neural operator demonstrates a better alignment with true modes than the neural operator alone.

\section{Discussion}
\label{sec:discussion}

\subsection*{Mitigating spectral bias in neural operators}

Trained neural operators can serve as fast surrogates for modeling physical systems when subjected to unseen initial or boundary conditions, without the need for solving the system again using traditional discretization-based numerical solvers.
However, the states inferred by the neural operator often tend to suffer from over-smoothening. 
In this study, we systematically compare the distribution of energy across the spectrum of Fourier modes, for the true solution and the states predicted by the neural operator. 
The energy of the state at any time step can be considered as the square of the absolute of the Fourier transform of the corresponding state. 
For all the cases reported in \autoref{sec: results}, the energy spectrum analysis of the states demonstrates that the energy of the solutions inferred by the neural operator aligns well with the ground truth only for lower wave numbers.
For higher wave numbers, the energy of neural operator prediction was consistently lower than that of the true solution. 
This behavior is called the spectral bias of neural operators and often becomes more prominent when the prediction window horizon is longer.  
The problem of oversmoothening arising from spectral bias of neural operator becomes a significant drawback while modeling turbulent flows using neural operators, because turbulent systems retains non-trivial energy even at higher wavenumber that cannot be approximated by the neural operators in the conventional setup.
Our results demonstrate that a diffusion model conditioned on a neural operator as its prior can overcome this issue and provide estimates of the state with a better aligned energy spectrum.

\subsection*{Computational Costs}

The diffusion model has to autoregressively perform $n$ denoising steps to estimate the state of the system at any time step as shown in \autoref{fig: diffusion_process}. 
Recent advances and improvements in the training and sampling of diffusion models \cite{karras2022elucidating} helped to bring down the number of sampling steps from $n=1000$ to $n=32$.
For the Kolmogorov flow test case, the trained neural operator can predict a trajectory from an unseen initial state in roughly $0.02s$, whereas, the diffusion model requires $0.9s$.
Although the diffusion model is slower than the neural operator at the inference, it is significantly faster than the traditional discretization-based solvers and achieves the true spectral properties unlike the neural operators.
All the neural operators and the diffusion models consist of $\approx 2M$ trainable parameters. 
The neural operators were trained using the cosine annealing based learning rate scheduler, and the diffusion models were trained with a constant learning rate of $10^{-4}$ as recommended in \cite{karras2022elucidating}. 
Our diffusion model was implemented based on \cite{karras2022elucidating, wang2020denoising}.

\subsection*{Impact of Neural Operator Accuracy on Diffusion Model}
\begin{figure}[h!]
  \centering
  \includegraphics[width=0.92\textwidth]{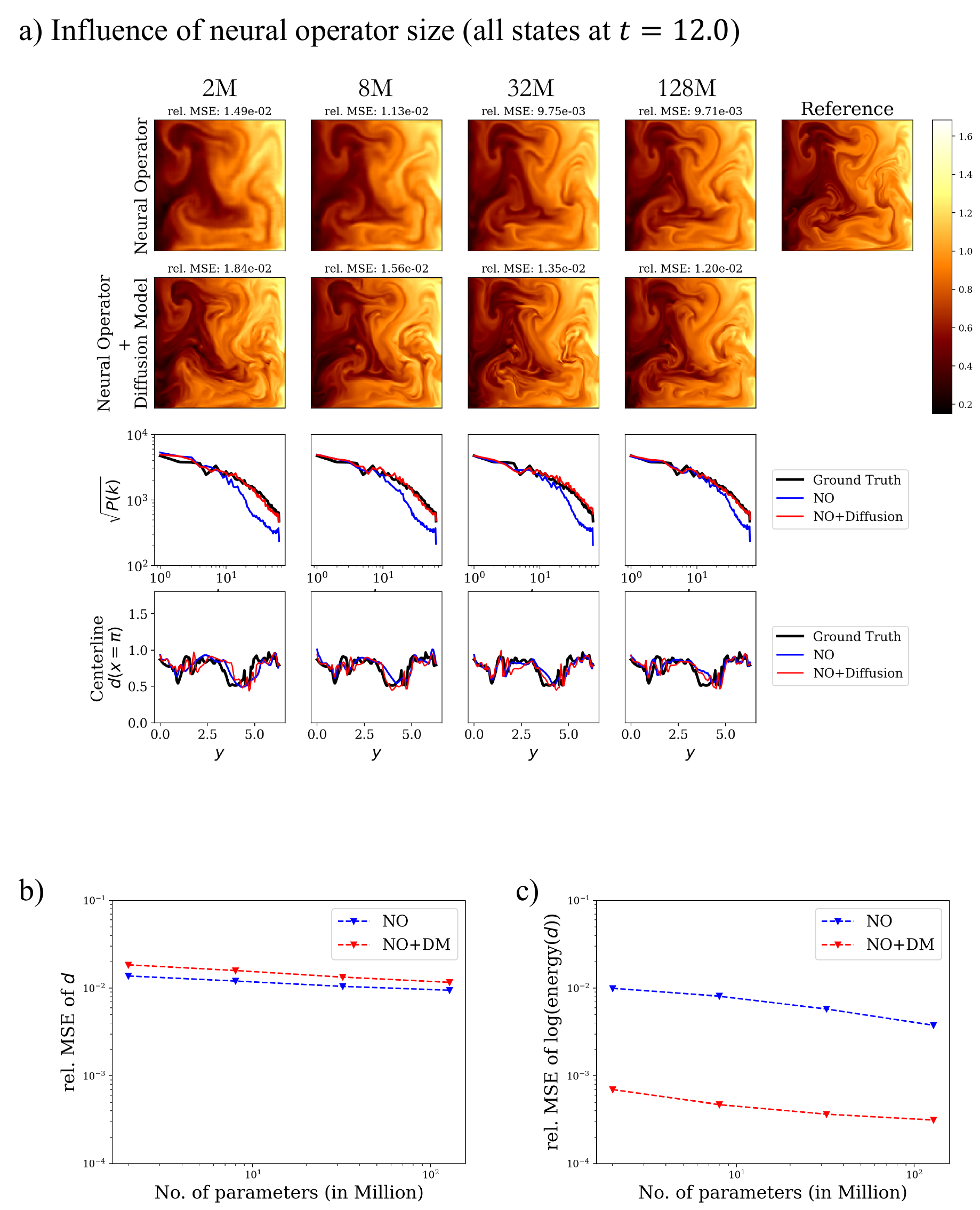}
  \caption{\textbf{Sensitivity to neural operator size.} We compare the states at $t=12$ for a sample from test dataset predicted by diffusion model conditioned on neural operators with 2M, 8M, 32M, and 128M parameters. The energy spectrum and centreline plots are also provided. b) and c) represents the change in relative MSE of the concentration field $d$ and relative MSE of the log spectrum of $d$.}
  \label{fig: buoyancy_analysis}
\end{figure}

A careful analysis of the buoyancy-driven flow results in \autoref{fig: buoyancy} reveals a discrepancy between the true and estimated states, especially at later time steps, due to the inability of the neural operator to estimate the high-frequency content that has low energy. 
At each autoregressive step this error cascades up the energy spectrum towards the low-frequency content that bear higher energy.
Although the diffusion model offers a better-aligned energy spectrum by recovering the missing frequencies, the error introduced by the neural operator does not go away. 
This motivated us to investigate the impact of neural operator accuracy on the diffusion model by varying the number of trainable parameters of neural operator as 2M, 8M, 32M, and 128M.
The diffusion model has only 2M parameters and is not varied.
We analyze the concentration field $d$ at $t=12$, predicted by all the neural operators and diffusion models in \autoref{fig: buoyancy_analysis}.
Increasing the number of trainable parameters enables the neural operator to better learn the high frequency content, subsequently leading to lower errors.
An improved neural operator also serves as a better prior to the diffusion model. 
The relative MSE of $d$ predicted by the neural operator alone and by the diffusion model conditioned on the neural operator is reduced by increasing the number of trainable parameters in the neural operator. 
However, the diffusion model's predictions have slightly higher mean  errors than the neural operator's output.
Although the diffusion model helps to recover the missing frequencies, they do not occur at all the correct locations resulting in slightly higher errors on the average, as shown in the centerline plots due to misalignment or phase errors. 
Regarding the energy spectrum, the diffusion model offers an order-of-magnitude improvement over the neural operator on the relative MSE.
To summarize, both the field and spectrum error of the diffusion model's predictions decreases if we have an improved prior from a more accurate neural operator. 
If there is a cost or memory constraint, for example, 4M trainable parameters, the diffusion model (2M) conditioned on the neural operator (2M) can still offer predictions with well-aligned energy spectrum. 

\subsection*{Is field MSE the perfect evaluation metric?}

We have added \autoref{tab:results} in order to address this question. We show two metrics - the field error and the energy spectrum error. The field error corresponds to the relative MSE between the ground truth and the prediction, and the energy spectrum error corresponds to the relative MSE between the log power spectrum of the ground truth and the prediction. 
The field error of NO (Neural Operator) is slightly better than NO+DM (Neural Operator + Diffusion Model). 
The energy spectrum error of NO+DM is better than NO. 
In order to understand this, we need to carefully examine what each metric represents. The field error does not account for all frequencies uniformly. 
It is more sensitive to those frequencies that carry higher energy, and is less impacted by the frequencies that bear lower energy.
In all the turbulent fluid flow problems considered in this study, the energy spectrum reveals that most of the energy is carried by the lower frequencies. 
Therefore, the field error metric dominantly represents the ability of the surrogate model to accurately reconstruct the mean flow represented by the lower frequencies. 
The high frequency small-scale structures that bear less energy are under-represented in the field error metric. 
As a consequence, the neural operator trained to minimize the field error naturally becomes agnostic to small scales, but in turbulent flows the small scales are important!
On the contrary, NO+DM can capture both the large and small-scale turbulent structures.
This indicates that the NO+DM model accurately captures the statistical structure of the turbulence compared to NO. Hence, higher field error and lower energy spectrum error indicate that the the underlying model respects the global statistical features and energy dynamics at the cost of some local phase alignment issues. Given this, we should also note that the field error of NO+DM is quite small in itself.

\begin{table}[h!]
\centering
\setlength{\arrayrulewidth}{0.3mm}
\setlength{\tabcolsep}{5pt}
\renewcommand{\arraystretch}{1.5}
\begin{tabular}{|>{\raggedright\arraybackslash}m{5.4cm}|>{\centering\arraybackslash}m{2cm}|>{\centering\arraybackslash}m{2cm}|>{\centering\arraybackslash}m{2cm}|>{\centering\arraybackslash}m{2cm}|}
\hline
\textbf{} &
  \multicolumn{2}{c|}{\textbf{Field Error}} &
  \multicolumn{2}{c|}{\textbf{Energy Spectrum Error}} \\ \hline
\textbf{} &
  \textcolor{blue}{\textbf{NO}} &
  \textcolor{red}{\textbf{NO+DM}} &
  \textcolor{blue}{\textbf{NO}} &
  \textcolor{red}{\textbf{NO+DM}} \\ \hline
Kolmogorov Flow - FNO &
  \textbf{\textcolor{blue}{5.94e-02}} &
  \textcolor{red}{6.53e-02} &
  \textcolor{blue}{9.09e-02} &
  \textbf{\textcolor{red}{1.97e-03}} \\ \hline
Kolmogorov Flow - UNet &
  \textbf{\textcolor{blue}{5.52e-02}} &
  \textcolor{red}{6.04e-02} &
  \textcolor{blue}{1.90e-02} &
  \textbf{\textcolor{red}{2.57e-03}} \\ \hline
Kolmogorov Flow - TC-UNet &
  \textbf{\textcolor{blue}{4.80e-02}} &
  \textcolor{red}{5.38e-02} &
  \textcolor{blue}{1.44e-02} &
  \textbf{\textcolor{red}{1.99e-03}} \\ \hline
Buoyancy-driven Flow &
  \textbf{\textcolor{blue}{1.20e-02}} &
  \textcolor{red}{1.58e-02} &
  \textcolor{blue}{8.10e-03} &
  \textbf{\textcolor{red}{4.71e-04}} \\ \hline
Turbulent Airfoil Wake &
  \textbf{\textcolor{blue}{8.35e-04}} &
  \textcolor{red}{1.25e-03} &
  \textcolor{blue}{8.82e-04} &
  \textbf{\textcolor{red}{7.64e-04}} \\ \hline
Turbulent Jet &
  \textbf{\textcolor{blue}{7.87e-03}} &
  \textcolor{red}{1.57e-02} &
  \textcolor{blue}{1.31e-02} &
  \textbf{\textcolor{red}{1.02e-03}} \\ \hline
\end{tabular}
\caption{{\color{blue}Comparison of Field Error and Energy Spectrum Error for different methods and cases.}}
\label{tab:results}
\end{table}

\subsection*{Diffusion model stabilizes autoregressive rollouts}
\begin{figure}[h!]
  \centering
  \includegraphics[width=0.9\textwidth]{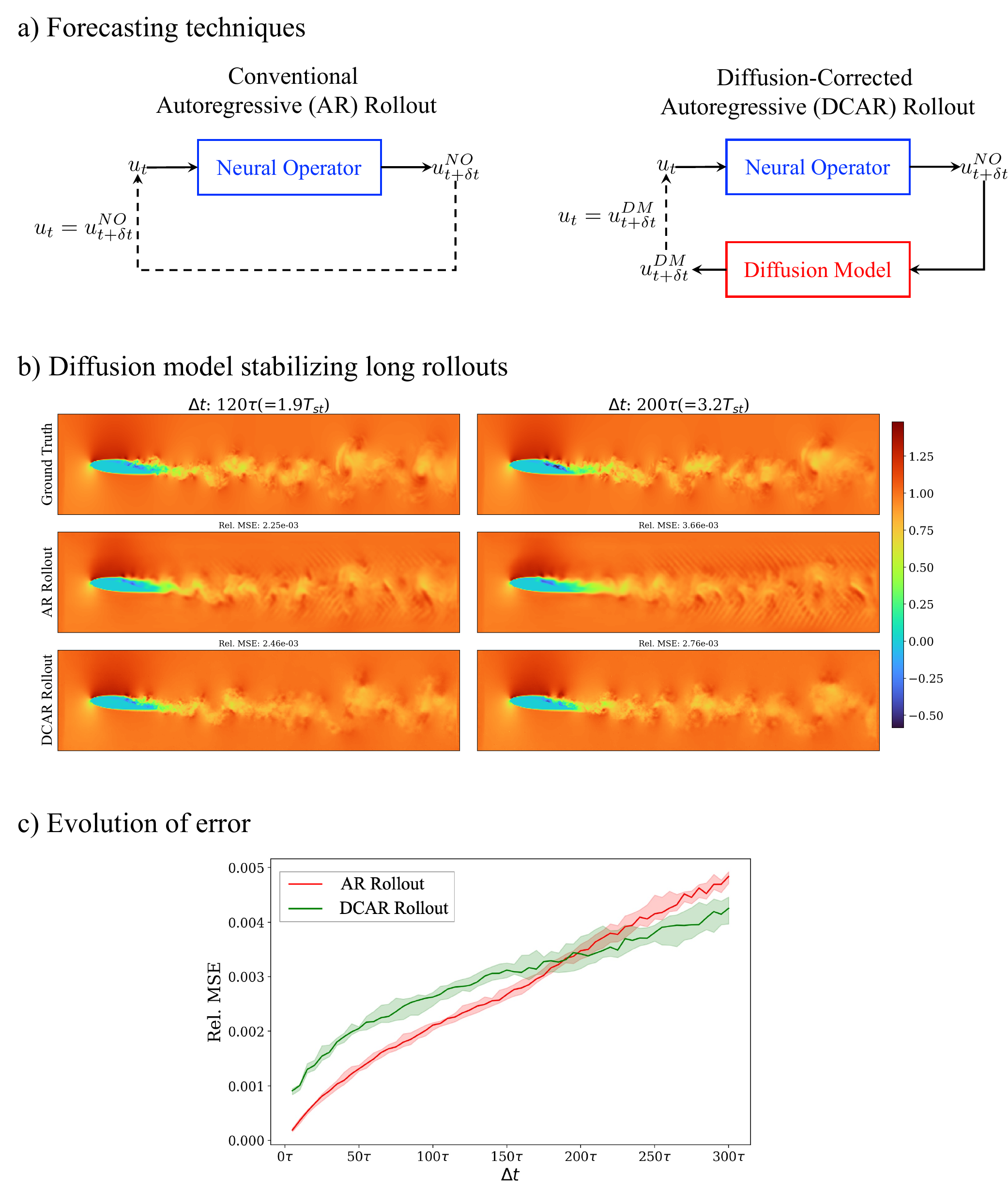}
  \caption{\textbf{Long Rollout.} a) Illustrates the conventional and diffusion-corrected autoregressive rollout strategies. b) Compares the velocity field at $\Delta t = $ 120$\tau$ and 200$\tau$ into the future predicted through autoregressive rollouts. (where $\tau$ represents the convective time scale). c) Represents the evolution of error during the autoregressive rollout, across all the samples in the test dataset. } 
  \label{fig: long_rollout}
\end{figure}

Although neural operators are computationally efficient, performing long-term forecasts requires autoregressive rollouts where the output of the neural operator is fed back as the input. 
Unfortunately, error accumulates at each autoregressive step making longer rollouts unstable. 
The spectral bias of neural operators further aggreviates the quality of autoregressive rollouts. 
Here we investigate if diffusion models that mitigate spectral bias in neural operators have the ability to stabilize autoregressive rollouts. 
To this end, we train a UNet-based neural operator, \cite{ovadia2023ditto}, that takes a history of states - $\{ u_{t-50\tau}, u_{t-40\tau}, ..., u_t \}$ as the input and directly predicts the states - $\{ u_{t+10\tau},  u_{t+20\tau} \}$. 
We  then auto-regressively predict the output upto $u_{t+300\tau}$. 
For the neural operator case, during the autoregressive (AR) rollout, we feed back the output predicted by the neural operator as its input to forecast further into the future.
In the diffusion-corrected autoregressive (DCAR) rollout, the output of the neural operator is corrected by the diffusion model to recover the missing frequencies, before feeding it back as the input to the neural operator. 
We show b) the contour plots of two snapshots at $ u_{t+120\tau}$ and $ u_{t+200\tau}$ and c) the evolution of relative MSE across the auto-regressive steps in Fig \ref{fig: long_rollout}. 
It can be seen that the NO (Neural Operator) prediction has some unrealistic artifacts, when predicting auto-regressively. 
The artifacts become more prominent as we forecast further into the future. 
On the other hand, the diffusion model helps stabilize longer rollouts into the future by reconstructing the missing frequencies at every autoregressive step. 
Moreover, it can be seen that initially the AR rollout predictions has lower Rel. MSE but as longer forecasts are carried out, at later time steps the Rel. MSE of the AR rollout exceeds that of the DCAR rollout.
It is interesting to observe that at $\Delta t = 120\tau $, the AR rollout with the unrealistic artifacts have a lower rel. MSE (2.25e-03) than DCAR rollout prediction (2.46e-03).
The unexpected observation occurs because these high-frequency artifacts bear low energy and remain less significant while computing the relative MSE of the velocity field predictions. 
This behavior further amplifies the significance of the question - Is field MSE the perfect evaluation metric for turbulent systems.

\subsection*{Proper Orthogonal Decomposition Analysis }

\begin{figure}
  \centering
  \includegraphics[width=0.65\textwidth]{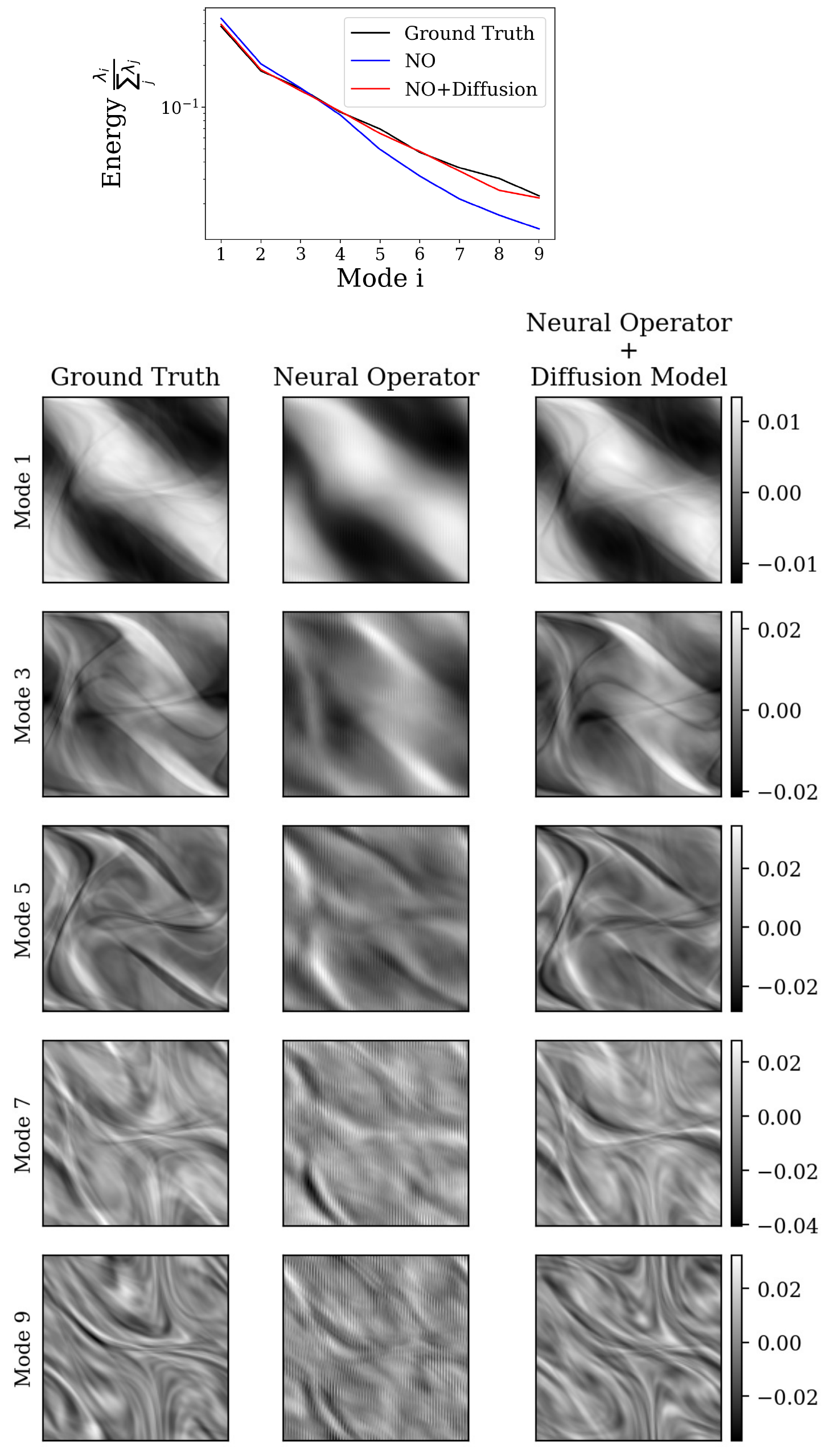}
  \caption{\textbf{POD analysis of Kolmogorov flow.} We compare the decay of eigenvalues obtained from performing proper orthogonal decomposition of 10 snapshots predicted by the neural operator (FNO) and diffusion model conditioned on the neural operator with the ground truth solution. Additionally, we also visualize and compare POD modes corresponding to $i=1,3,5,7,9$.}
  \label{fig: pod_kolmogorov}
\end{figure}

The energy spectra in \autoref{fig: kolmogorov} to \autoref{fig: schlieren} demonstrate that the diffusion model conditioned on the neural operator performs better than the neural operator alone. 
However, these plots are constructed for each time step separately and do not contain information about the cross-correlation of the states across time steps. 
We address this question by performing proper orthogonal decomposition (POD) on 10 timesteps predicted by the neural operator, as well as the diffusion model conditioned on the neural operator, and compare it with the corresponding state of the ground truth.
\autoref{fig: pod_kolmogorov} illustrates the decay of the eigenvalues and compares the POD modes of the true data with the other two approaches for the Kolmogorov flow case.
The proposed framework of the diffusion models conditioned on the neural operator has better aligned eigen spectrum and eigen modes with the ground truth than the neural operators alone. 
The POD analysis of buoyancy-driven flow (\autoref{fig: pod_buoyancy}), turbulent airfoil (\autoref{fig: pod_airfoil}), 3D turbulent jet (\autoref{fig: pod_jet_3d}) and Schlieren velocimetry of helium jet (\autoref{fig: pod_schlieren}) are also provided in \autoref{app: pod_analysis}.

\subsection*{Future Directions}

Summarizing, we propose a framework that utilizes neural operators as effective priors for diffusion models. 
Our results demonstrate that the proposed framework that blends neural operators with diffusion models significantly improves the energy spectrum compared to the neural operators alone and has been tested across different types of neural operator architectures - FNO, U-Net, and TC-UNet, over a variety of test cases. 
The framework utilized here is general, and therefore, any advances in the quality, computational efficiency, or scalability of either the neural operator or the diffusion models in the future can be applied seamlessly and could potentially improve the proposed framework.

The sampling routine of the score-based diffusion model we adopted consists of 32 denoising steps, corresponding to 32 function evaluations as against 1 function evaluation for the neural operator. 
Therefore, the diffusion model is approximately 32 times slower than the neural operator at inference.
Nevertheless, inferring states of the system using diffusion models still takes only an order of $~1 s$ on an NVIDIA H100 GPU, for all the cases we considered.  
Improving the accuracy of longer rollouts, incorporating conservation of physical laws, and applying the methodology to problems in other scientific domains are interesting future research directions. 
We also believe it is worth investigating the potential of combining neural operators with other generative modeling techniques like GANs and normalizing flows.

The utility of this new method is in accelerating DNS and LES by interweaving small bursts of such expensive simulations with large extrapolation time intervals from our method. This is similar to previous work we did for materials problem \cite{oommen2024rethinking} using the neural operators alone. Moreover, for experimental work where a few snapshots are visualized, the present method will enable a continuous in time
evolution of the turbulent field at DNS-level fidelity.

\section{Methods}
\label{sec: Methods}

\subsection{Neural Operators} 
Neural Operators are a class of architectures in deep learning, that maps between infinite dimensional functional spaces. Consider an operator $\mathcal{G}$, that maps from the input function space $U$ to the output function space $V$, i.e., $\mathcal{G}: U \rightarrow V$. Then any machine learning architecture that is used to approximate this functional mapping to predict $v \in V$ for a given $u \in U$ can be defined as a neural operator. Neural operator is a popular framework for encoding dynamics and making predictions under some given conditions. But like any other neural network architectures, they suffer from spectral bias. 

\textbf{Lemma 4.1.} Let  $F:U \rightarrow V$ be a functional mapping with $U,V \in H^{s}(\Omega)$, where $H^{s}(\Omega)$ is the Sobelov space of functions defined on $\Omega$. Let $\mathcal{G}(\theta_{NO})$ be a neural operator parameterized by $\theta_{NO}$, that is approximating the functional space $F$. Then $\mathcal{G}$ learns the lower frequency components of the functional space $F$ more effectively \cite{rahaman2019spectral,cao2019towards, bowman2023spectral,basri2020frequency}.

\textbf{Sketch of Proof:} We want to approximate the functional space $F(U) = V$, then if we take the Fourier transform for this function, we have
\begin{equation}
    \hat{F}(\xi) =\int_{-\infty}^{\infty} F(U)e^{-2i\pi \xi U}dU,
\end{equation} and\begin{equation}
    F(U) = \int_{-\infty}^{\infty}\hat{F}(\xi)e^{2i\pi \xi U}d\xi,
\end{equation} where $ \hat{F}(\xi)$ represents the amplitude of the the frequency component $\xi$. Now let $\mathcal{G}(\theta_{NO}; U)$ is the operator approximating $F(U)$, typically the loss function $\mathcal{L}(\theta_{NO})$ used for optimizing such networks is given by \begin{equation}
    \mathcal{L}(\theta_{NO}) = \frac{1}{N}\sum_{n=1}^{N}(F(u_{n})-\mathcal{G}(\theta_{NO};u_{n}))^{2},
\end{equation}where $u_{n}\in U$,  and $n=1,2,\cdot \cdot \cdot , N$ are the discrete points where the functional evaluation are available for training. We can also express $\mathcal{G}$ in terms of its discrete Fourier components as
\begin{equation}
    \hat{\mathcal{G}}(\theta_{NO}; \xi) =\frac{1}{N}\sum_{n=1}^{N-1} \mathcal{G}(\theta_{NO}; u_{n})e^{-2\pi i\xi u_n/ N},
\end{equation}and\begin{equation}
    \mathcal{G}(\theta_{NO}; u_{n}) = \frac{1}{N}\sum_{n=1}^{N-1}\hat{\mathcal{G}}(\theta_{NO}; \xi)e^{2\pi i\xi u_n/N}.
\end{equation} During the minimization of the loss functions the parameters are updated based on the gradient descent update rule, given by \begin{equation}
    \theta_{NO}^{(m+1)} = \theta_{NO}^{(m)} - \alpha\nabla_{\theta_{NO}}\mathcal{L}(\theta_{NO}^{(m)}), \hspace{0.2in} m = 1,2, \cdot \cdot \cdot , M
\end{equation} where $M$ is the total number of iterations, $\alpha$ is the learning rate and $\nabla_{\theta_{NO}}\mathcal{L}$ is given by \begin{equation}
    \nabla_{\theta_{NO}}\mathcal{L}(\theta_{NO}) = \frac{-2}{N}\sum_{n=1}^{N}(F(u_{n})- \mathcal{G}(\theta_{NO}; u_{n}))\nabla_{\theta_{NO}}\mathcal{G}(\theta_{NO}; u_{n}).
\end{equation}Hence, mathematically the rate of change of the Fourier coefficients $\hat{\mathcal{G}}(\theta_{NO}; \xi)$ with respect to training iteration $m$ can be given by \begin{equation}
    \frac{d}{dm}(\hat{\mathcal{G}}(\theta_{NO}; \xi)) \propto -\alpha\hat{\nabla}_{\theta_{NO}}\mathcal{L}(\theta_{NO}) \propto \frac{\partial \mathcal{G}(\theta_{NO}; \xi)}{\partial \theta_{NO}}.
\end{equation} Since typically in neural network architectures $\partial \mathcal{G}(\theta_{NO}; \xi)/\theta_{NO}$ is smooth or piecewise smooth, $\alpha\hat{\nabla}_{\theta_{NO}}\mathcal{L}(\theta_{NO})$ is larger for smaller $\xi$ \cite{rahaman2019spectral,basri2020frequency,stein1971introduction}. Therefore, the operator $\mathcal{G}$ learns the low frequency components of the functional space $F(U)$ more effectively compared to higher frequency components, which is known as the spectral bias. Hence like any other networks, operators suffer from spectral bias, which leads to ineffective refinement of the higher frequency component, leading, in turn, to sharp decay of the power spectrum associated with the output predicted by any neural operator. Although neural operators fail to capture the high frequency, it captures effectively the mean flow dominated by the low wave numbers. This acts as an ideal prior for diffusion models to mimic the true flow consisting of all the frequencies. 

\subsection{Score-based diffusion models} 

Diffusion models \cite{ho2020denoising} are generative algorithms that can effectively generate samples from the true underlying distribution of functions, $\mathcal{T}$, given a finite set of samples from $\mathcal{T}$. 
The diffusion models generate samples from $\mathcal{T}$ by learning to travel from a simple distribution such as a standard normal distribution ($\Gamma_0 =\mathcal{N}(0,I)$) to the desired complex distribution ($\Gamma_N \approx \mathcal{T}$) through $N$ iterative steps. 
Specifically, at first the diffusion model takes the function $\mathbf{X_0}$ sampled from $\Gamma_0$ as the input and predicts $\mathbf{X}_1$. 
Since $\Gamma_0 = \mathcal{N}(0,I)$, $\mathbf{X}_0$ can be easily sampled. 
Now, the diffusion model can estimate $\mathbf{X}_{i+1}$ from $\mathbf{X}_i$ iteratively for $N$ steps. 
But how can we train the diffusion model to iteratively transition from $\Gamma_0 =\mathcal{N}(0,I)$ to $\Gamma_N \approx \mathcal{T}$, the desired true distribution, when we do not have ground truth samples from the intermediate distributions $\Gamma_i$ for $i=\{1,2,3, \cdots, n-1\}$? 
This can be achieved through denoising score matching with Langevin dynamics \cite{song2020score}.

The overall goal of a score ($s$) based diffusion model is to estimate the score function defined by $s_{\theta_{D}}(\mathbf{X})=\nabla_{X}\log p(\mathbf{X})$, where $\theta_{D}$ are the parameters of the diffusion model and $p$ is the probability density of $\mathbf{X}$, where $\mathbf{X}$ is the continuous representation of $\mathbf{X}_i \sim \Gamma_i$.
Since the data distribution is not known explicitly, and there is a chance that the data might lie on a lower dimensional manifold, the score function may be ill defined in regions where there is no data. To circumvent this issue, perturbation of the data is done by adding Gaussian noise, which ensures that the score is well defined across the entire space by smoothing the distribution. Hence, the score function gives us a direction to move towards the region of higher probability. However the direct way to sample from the distribution is missing here. In order to fix this problem, Langevin Dynamics was used in \cite{song2019generative}. Langevin Dynamics ensures the convergence of the generated samples to the true underlying distribution. It balances the inherent dynamics between deterministic movement, which is driven by the gradient of log of the probability, with random exploration given by the noise levels. In our proposed method, we use the score function, which is additionally conditioned on the output of the neural operator $\mathbf{\mathcal{G}}(\theta_{NO}; (\mathbf{x,t}))$ (for readability expressed as $\mathcal{G}$) given by
\begin{equation} 
\label{eq1d}  
s_{\theta_{D}}(\mathbf{X},\sigma,\mathbf{\mathcal{G}}) = \nabla_{X}\log p(\mathbf{X}|\mathbf{\mathcal{G}}),
\end{equation} 
where $\sigma$ is the noise level. The modified score function now directs the diffusion model to sample from the posterior distribution of $\mathbf{X}$ given $\mathbf{\mathcal{G}}$. This ensures that the the generated samples from the diffusion model are consistent with both the structures defined by $\mathbf{\mathcal{G}}$ and the data distribution that is learned by the diffusion model. The update rule for Langevin dynamics can be shown as
\begin{equation} \label{eq2d}
    \mathbf{X}_{i+1} = \mathbf{X}_{i} + \frac{\varepsilon}{2}s_{\theta_{D}}(\mathbf{X}_{i},\sigma_{j},\mathbf{\mathcal{G}})+\sqrt{\varepsilon}z_{i},
\end{equation}where $\varepsilon$ is the step size, $z_{i}$ is the noise component and $\sigma_{j}$ is the scale of noise at the particular time step $j$ during the diffusion process. From Eqns \ref{eq1d} and \ref{eq2d}, we can see that as $\sigma_{i}$ decreases, the score function $s_{\theta_{D}}$ focuses on recovering the high-frequency details which was smoothen out by the noise. In the frequency domain Eqn \ref{eq2d} can be shown as \begin{equation}\label{lvs}
    \hat{\mathbf{X}}_{i+1} = \hat{\mathbf{X}}_{i} +  \frac{\varepsilon}{2}\hat{s}_{\theta_{D}}(\mathbf{X}_{i},\sigma_{j},\mathbf{\mathcal{G}})+\sqrt{\varepsilon}\hat{z}_{i},
\end{equation}where $\hat{\mathbf{X}}_{i}$, $\hat{s}$ and $\hat{z}$ are the Fourier transform of the respective functions. Hence, after multiple iterations as the noise diminishes $\hat{z}$ becomes negligible and the score function amplifies the high frequency component as it acts as a high pass filter in reverse.

\textbf{Lemma 4.2.} Let $u(\mathbf{x,t})$ $\in$ $H^{s}(\Omega):\{u \in L^{2}(\Omega)|\partial^{\alpha}u \in L^{2}(\Omega), \forall |\alpha|\leq s\}$ be the true solution discretized over N spatial grid points.
\begin{equation}
    u(\mathbf{x,t}) = \sum_{n=1}^{\infty}a_{n}(\mathbf{t})\phi_{n}(\mathbf{x}),
\end{equation}
where $\phi_{n}(\mathbf{x})$ are the basis functions and $a_{n}(\mathbf{t})$ are the time-dependent coefficients. 
\newline 
Let $\mathbf{\mathcal{G}}$ and $s_{\theta_D}(\mathbf{X},\mathbf{\mathcal{G}},\sigma)$ represents the output of the neural operator and the diffusion model respectively, such that 
\begin{equation}
    s_{\theta_D}(\mathbf{X},\mathbf{\mathcal{G}},\sigma) \simeq  \sum_{n=1}^{\infty}a_{n}^s(\mathbf{t})\phi_{n}(\mathbf{x}).
\end{equation}


The spectral bias suffered by neural operators limits its ability to learn the true solutions beyond a critical wavenumber $|k_c|$.
Let $N_c$ represent the critical basis index in the spectrally decomposed representation of $\mathcal{G}$ beyond which the neural operator's prediction carries trivial amount of energy. In this context, $|k_c|$ can interpreted as the wavenumber that carries maximum energy in the true mode $\hat{\phi}_{N_c}(k)$. The energy of $\mathcal{G}$, $E_{\mathcal{G}}(t, k)$ can now be represented as,
\begin{equation}
    E_{\mathcal{G}}(t, k) = \frac{1}{N}\left| \sum_{n=1}^N a_n^{\mathcal{G}}(t) \hat{\phi}_n(k) \right|^2 \approx \frac{1}{N} \left| \sum_{n=1}^{N_c} a_n^{\mathcal{G}}(t) \hat{\phi}_n(k) \right|^2,
\end{equation}
because  $ \left| \sum_{n=N_c+1}^{N} a_n^{\mathcal{G}}(t) \hat{\phi}_n(k) \right|^2 \approx 0$.
From \autoref{fig: kolmogorov} we see that using a better neural operator has a better aligned energy spectrum implying a larger $k_c$. 
Subsequently, the output of the diffusion model also improves as shown in \autoref{tab:results}.  
In other words, a better $\mathcal{G}$ serves as an improved prior increasing $k_c$ and $\left| u(\mathbf{x},t)-s_{\theta_D}(\mathbf{x},\mathbf{\mathcal{G}},\sigma)) \right|\rightarrow 0$.

\textbf{Sketch of Proof:} 
We have, 
\begin{align}
    s_{\theta_D}(\mathbf{X},\mathbf{\mathcal{G}}, \sigma)  & \simeq  \sum_{n=1}^{\infty}a_{n}^s(t)\phi_{n}(\mathbf{X}) \\
      & = \sum_{n:|k|\leq |k_{c}|}a_{n}^s(t)\phi_{n}(\mathbf{x}) +  \sum_{n:|k|> |k_{c}|}a_{n}^s(t)\phi_{n}(\mathbf{x}).
\end{align}
Let us assume, in the discrete Fourier space we can write it in terms of the energy spectrum as
\begin{equation}
    E_{s_{\theta_D}}(k) = \frac{1}{N} \left| \sum_{ (n\leq N_c)}a_{n}^s(t)\hat{\phi_{n}}(k)\right|^{2} + \frac{1}{N} \left| \sum_{ (n > N_c)}a_{n}^s(t)\hat{\phi_{n}}(k)\right|^{2},
\end{equation}
From Eqn \ref{eq1d} in the frequency domain, the score function can be written as with the assumption that the  conditional distribution $p$ follows a Gaussian distribution  \begin{equation}
    \hat{s}(\mathbf{X},\mathbf{\mathcal{G}},\sigma_{j}) \simeq \nabla_{\hat{X}}\log p (\hat{\mathbf{X}}(k)|\hat{\mathcal{G}}(k)) = -\frac{\hat{\mathbf{X}}(k) - \hat{\mathcal{G}}(k)}{var(k,\sigma_{j})},
\end{equation}
where $\hat{\mathbf{X}}(k)$ is the Fourier transform of the current state at wave number $k$ and $var(k,\sigma_{j})$ is the variance of the distribution conditioned on the noise level $\sigma_{j}$ at wave number $k$. Although the specific relation between $var(k,\sigma_{j})$ and $k$ is problem specific, typically \cite{proakis2007digital}, 
\begin{equation}
    var(k,\sigma_{j}) \propto \frac{1}{k^{p}} + \sigma_{j}^{2},
\end{equation} where $p$ is a positive constant based on the data. Hence, as $k \rightarrow 0$,  $var(k,\sigma_{j})$ increases and $ \frac{\hat{u}(k) - \hat{\mathcal{G}}(k)}{var(k,\sigma_{j})}\rightarrow 0$. This implies a weak gradient value signifying smaller steps. Now if we look at the frequency domain of the Langevin dynamics, from Eqn \ref{lvs} \begin{equation}
    \hat{\mathbf{X}}_{j+1} = \hat{\mathbf{X}}_{j}(k) -\frac{\varepsilon}{2}\frac{\hat{\mathbf{X}}(k) - \hat{\mathcal{G}}(k)}{var(k,\sigma_{j})}+\sqrt{\varepsilon}\hat{z}_{j}(k), 
\end{equation}
it can be seen that for smaller $k$ the above equation changes to \begin{equation}
    \hat{\mathbf{X}}_{j+1} \simeq \hat{\mathbf{X}_{j}}(k)+\sqrt{\varepsilon}\hat{z}_{j}(k).
\end{equation} 
This signifies that the Langevin dynamics updates  are dominated by the noise term $\sqrt{\varepsilon}\hat{z}_{j}(k)$ instead of the gradient driven term and hence leads to ineffective refinement of the low wave numbers. Hence, in an intermediate step during the diffusion process when the updates are dominated by the noise term, the energy spectrum can be given as \begin{equation}
    E_{s_{\theta_D},j+1}(k) = \frac{|\hat{\mathbf{X}}_{j}(k)|^{2}}{N} + \frac{\varepsilon|\hat{z}_{j}(k)|^{2}}{N} + \frac{2\sqrt{\varepsilon}\Xi[\hat{\mathbf{X}}_{j}(k)\hat{z}^{\ast}_{j}(k)]}{N},
\end{equation} 
where $\Xi[.]$ represents the real part of the complex number and $\hat{z}^{\ast}$ the complex conjugate of $\hat{z}$. Due to weaker gradients, the term  $|\hat{\mathbf{X}}_{j}(k)|^{2}$ grows slowly and $\varepsilon|\hat{z}_{j}(k)|^{2}$ may dominate, which might lead to flatter energy spectrum at low wave number and may not accurately represent the large scale features. 

Therefore, as as $k_{c}$ increases $\left| u(\mathbf{x},t)-s_{\theta_D}(\mathbf{X},\mathbf{\mathcal{G}},\sigma)) \right|\rightarrow 0$, or in other words for the proposed method to work effectively, the $\mathbf{\mathcal{G}}$ should be at-least be able to capture the mean flow, in order for the diffusion model to be effectively helping in adding the high frequency features to the predicted mean flow.

In summary, based on \textbf{Lemma 4.1.} and \textbf{Lemma 4.2.} we propose our method where we use a neural operator as a prior to score-based diffusion model. In step 1 we train a neural operator that accurately captures the low frequency wave numbers and hence the mean flow dynamics. In step 2 we pass the neural operator predicted outputs to the diffusion model, which acts a prior, which we further train based on the score function that focuses on the joint distribution and the high frequency components are added which leads to the recovery of the full spectrum. We note that as shown in \textbf{Lemma 4.2.}, the neural operator should be able to capture the mean flow accurately for the proposed method to work effectively.


\newpage
\section*{Acknowledgements}
V.O. acknowledges Alan John Varghese, and Dr. Andreas Euan Robertson for their insightful suggestions and valuable discussions at various stages of this project.

\noindent \textbf{Funding:} 
This work was supported by the MURI grant (FA9550-20-1-0358), the ONR Vannevar Bush Faculty Fellowship (N0001422-1-2795)and the U.S. Department of Energy, Advanced Scientific Computing Research program, under the Scalable, Efficient and Accelerated Causal Reasoning Operators, Graphs and Spikes for Earth and Embedded Systems (SEA-CROGS) project, (DE-SC0023191).

\noindent \textbf{Author Contributions:} 
V.O.: 
Conceptualization, Methodology, Software, Formal Analysis, Investigation, Data Curation, Writing - Original Draft, Writing - Review \& Editing and Visualization.
A.B.:
Conceptualization, Methodology, Software, Formal Analysis, Investigation, Writing - Original Draft and Writing - Review \& Editing
Z.Z.:
Software, Validation, Formal Analysis, Investigation, Visualization, Writing - Original Draft and Writing - Review \& Editing
G.E.K.:
Validation, Formal Analysis, Resources, Writing - Review \& Editing, Supervision, Project administration and Funding acquisition

\noindent \textbf{Competing Interests:} 
The authors declare that they have no competing interests. 

\noindent \textbf{Data and Materials Availability:} 
The data for Kolmogorov flow was simulated using the publicly available pseudo-spectral solver provided in  \href{https://github.com/neuraloperator/physics_informed/blob/d1835d1e6ee9d7969455ceb36040389a23d04d85/solver/legacy_solver.py#L68}{FNO's repository}.
The buoyancy-driven transport data was from \href{https://github.com/pdearena/pdearena}{PDEArena}.
The LES datasets for turbulent airfoil and jet were downloaded from \href{https://deepblue.lib.umich.edu/data/collections/kk91fk98z?locale=en}{link}.
The Schlieren velocimetry dataset can be found  \href{https://zenodo.org/records/6136052}{here}.
The code developed in this study will be made available after publication in our GitHub repository - \url{https://github.com/vivekoommen/NeuralOperator_DiffusionModel}.

\newpage
\bibliographystyle{unsrt}  
\bibliography{references}  

\begin{thebibliography}{10}

\bibitem{kim1987turbulence}
John Kim, Parviz Moin, and Robert Moser.
\newblock Turbulence statistics in fully developed channel flow at low {R}eynolds number.
\newblock {\em Journal of fluid mechanics}, 177:133--166, 1987.

\bibitem{choi1994active}
Haecheon Choi, Parviz Moin, and John Kim.
\newblock Active turbulence control for drag reduction in wall-bounded flows.
\newblock {\em Journal of Fluid Mechanics}, 262:75--110, 1994.

\bibitem{du2000suppressing}
Yiqing Du and George~Em Karniadakis.
\newblock Suppressing wall turbulence by means of a transverse traveling wave.
\newblock {\em Science}, 288(5469):1230--1234, 2000.

\bibitem{raissi2019physics}
Maziar Raissi, Paris Perdikaris, and George~E Karniadakis.
\newblock Physics-informed neural networks: A deep learning framework for solving forward and inverse problems involving nonlinear partial differential equations.
\newblock {\em Journal of Computational Physics}, 378:686--707, 2019.

\bibitem{karniadakis2021physics}
George~Em Karniadakis, Ioannis~G Kevrekidis, Lu~Lu, Paris Perdikaris, Sifan Wang, and Liu Yang.
\newblock Physics-informed machine learning.
\newblock {\em Nature Reviews Physics}, 3(6):422--440, 2021.

\bibitem{toscano2024pinns}
Juan~Diego Toscano, Vivek Oommen, Alan~John Varghese, Zongren Zou, Nazanin~Ahmadi Daryakenari, Chenxi Wu, and George~Em Karniadakis.
\newblock From pinns to pikans: Recent advances in physics-informed machine learning.
\newblock {\em arXiv preprint arXiv:2410.13228}, 2024.

\bibitem{kiyani2023framework}
Elham Kiyani, Khemraj Shukla, George~Em Karniadakis, and Mikko Karttunen.
\newblock A framework based on symbolic regression coupled with extended physics-informed neural networks for gray-box learning of equations of motion from data.
\newblock {\em arXiv preprint arXiv:2305.10706}, 2023.

\bibitem{zhang2024discovering}
Zhen Zhang, Zongren Zou, Ellen Kuhl, and George~Em Karniadakis.
\newblock Discovering a reaction--diffusion model for {A}lzheimer’s disease by combining {PINNs} with symbolic regression.
\newblock {\em Computer Methods in Applied Mechanics and Engineering}, 419:116647, 2024.

\bibitem{ahmadi2024ai}
Nazanin Ahmadi~Daryakenari, Mario De~Florio, Khemraj Shukla, and George~Em Karniadakis.
\newblock {AI-A}ristotle: {A} physics-informed framework for systems biology gray-box identification.
\newblock {\em PLOS Computational Biology}, 20(3):e1011916, 2024.

\bibitem{de2023ai}
Mario De~Florio, Ioannis~G Kevrekidis, and George~Em Karniadakis.
\newblock {AI-L}orenz: {A} physics-data-driven framework for black-box and gray-box identification of chaotic systems with symbolic regression.
\newblock {\em arXiv preprint arXiv:2312.14237}, 2023.

\bibitem{cavity2024physics}
Eric Fowler, Christopher~J McDevitt, and Subrata Roy.
\newblock Physics-informed neural network simulation of thermal cavity flow.
\newblock {\em Scientific Reports}, 14(1):15203, 2024.

\bibitem{Fangying-GK}
Fangying Song and George~Em Karniadakis.
\newblock Variable-order fractional models for wall-bounded turbulent flows.
\newblock {\em Entropy}, 23(6):782, 2021.

\bibitem{lu2021learning}
Lu~Lu, Pengzhan Jin, Guofei Pang, Zhongqiang Zhang, and George~Em Karniadakis.
\newblock Learning nonlinear operators via {D}eep{ON}et based on the universal approximation theorem of operators.
\newblock {\em Nature Machine Intelligence}, 3(3):218--229, 2021.

\bibitem{li2020fourier}
Zongyi Li, Nikola Kovachki, Kamyar Azizzadenesheli, Burigede Liu, Kaushik Bhattacharya, Andrew Stuart, and Anima Anandkumar.
\newblock Fourier {N}eural {O}perator for parametric partial differential equations.
\newblock {\em arXiv preprint arXiv:2010.08895}, 2020.

\bibitem{raonic2023convolutional}
Bogdan Raonic, Roberto Molinaro, Tobias Rohner, Siddhartha Mishra, and Emmanuel de~Bezenac.
\newblock Convolutional neural operators.
\newblock In {\em ICLR 2023 Workshop on Physics for Machine Learning}, 2023.

\bibitem{li2022transformer}
Zijie Li, Kazem Meidani, and Amir~Barati Farimani.
\newblock Transformer for partial differential equations' operator learning.
\newblock {\em arXiv preprint arXiv:2205.13671}, 2022.

\bibitem{ovadia2023ditto}
Oded Ovadia, Eli Turkel, Adar Kahana, and George~Em Karniadakis.
\newblock {DiTTO}: {D}iffusion-inspired {T}emporal {T}ransformer {O}perator.
\newblock {\em arXiv preprint arXiv:2307.09072}, 2023.

\bibitem{sharma2024graph}
Anuj Sharma, Sukhdeep Singh, and S~Ratna.
\newblock Graph neural network operators: a review.
\newblock {\em Multimedia Tools and Applications}, 83(8):23413--23436, 2024.

\bibitem{oommen2024rethinking}
Vivek Oommen, Khemraj Shukla, Saaketh Desai, R{\'e}mi Dingreville, and George~Em Karniadakis.
\newblock Rethinking materials simulations: Blending direct numerical simulations with neural operators.
\newblock {\em npj Computational Materials}, 10(1):145, 2024.

\bibitem{bora2023learning}
Aniruddha Bora, Khemraj Shukla, Shixuan Zhang, Bryce Harrop, Ruby Leung, and George~Em Karniadakis.
\newblock Learning bias corrections for climate models using deep neural operators.
\newblock {\em arXiv preprint arXiv:2302.03173}, 2023.

\bibitem{oommen2022learning}
Vivek Oommen, Khemraj Shukla, Somdatta Goswami, R{\'e}mi Dingreville, and George~Em Karniadakis.
\newblock Learning two-phase microstructure evolution using neural operators and autoencoder architectures.
\newblock {\em npj Computational Materials}, 8(1):190, 2022.

\bibitem{peyvan2024riemannonets}
Ahmad Peyvan, Vivek Oommen, Ameya~D Jagtap, and George~Em Karniadakis.
\newblock Riemann{ON}ets: Interpretable neural operators for {R}iemann problems.
\newblock {\em Computer Methods in Applied Mechanics and Engineering}, 426:116996, 2024.

\bibitem{hao2023gnot}
Zhongkai Hao, Zhengyi Wang, Hang Su, Chengyang Ying, Yinpeng Dong, Songming Liu, Ze~Cheng, Jian Song, and Jun Zhu.
\newblock Gnot: {A} general neural operator transformer for operator learning.
\newblock In {\em International Conference on Machine Learning}, pages 12556--12569. PMLR, 2023.

\bibitem{zappala2024learning}
Emanuele Zappala, Antonio Henrique de~Oliveira Fonseca, Josue~Ortega Caro, Andrew~Henry Moberly, Michael~James Higley, Jessica Cardin, and David~van Dijk.
\newblock Learning integral operators via neural integral equations.
\newblock {\em Nature Machine Intelligence}, pages 1--17, 2024.

\bibitem{ye2024locality}
Ximeng Ye, Hongyu Li, Jingjie Huang, and Guoliang Qin.
\newblock On the locality of local neural operator in learning fluid dynamics.
\newblock {\em Computer Methods in Applied Mechanics and Engineering}, 427:117035, 2024.

\bibitem{wan2025deepvivonet}
Ruyin Wan, Ehsan Kharazmi, Michael~S Triantafyllou, and George~Em Karniadakis.
\newblock Deepvivonet: Using deep neural operators to optimize sensor locations with application to vortex-induced vibrations.
\newblock {\em arXiv preprint arXiv:2501.04105}, 2025.

\bibitem{xu2025understanding}
Zhi-Qin~John Xu, Lulu Zhang, and Wei Cai.
\newblock On understanding and overcoming spectral biases of deep neural network learning methods for solving pdes.
\newblock {\em arXiv preprint arXiv:2501.09987}, 2025.

\bibitem{cai2019multi}
Wei Cai and Zhi-Qin~John Xu.
\newblock Multi-scale deep neural networks for solving high dimensional pdes.
\newblock {\em arXiv preprint arXiv:1910.11710}, 2019.

\bibitem{cai2020phase}
Wei Cai, Xiaoguang Li, and Lizuo Liu.
\newblock A phase shift deep neural network for high frequency approximation and wave problems.
\newblock {\em SIAM Journal on Scientific Computing}, 42(5):A3285--A3312, 2020.

\bibitem{tancik2020fourier}
Matthew Tancik, Pratul Srinivasan, Ben Mildenhall, Sara Fridovich-Keil, Nithin Raghavan, Utkarsh Singhal, Ravi Ramamoorthi, Jonathan Barron, and Ren Ng.
\newblock Fourier features let networks learn high frequency functions in low dimensional domains.
\newblock {\em Advances in neural information processing systems}, 33:7537--7547, 2020.

\bibitem{wang2021eigenvector}
Sifan Wang, Hanwen Wang, and Paris Perdikaris.
\newblock On the eigenvector bias of fourier feature networks: From regression to solving multi-scale pdes with physics-informed neural networks.
\newblock {\em Computer Methods in Applied Mechanics and Engineering}, 384:113938, 2021.

\bibitem{lippe2024pde}
Phillip Lippe, Bas Veeling, Paris Perdikaris, Richard Turner, and Johannes Brandstetter.
\newblock Pde-refiner: Achieving accurate long rollouts with neural pde solvers.
\newblock {\em Advances in Neural Information Processing Systems}, 36, 2024.

\bibitem{zhang2022hybrid}
Enrui Zhang, Adar Kahana, Eli Turkel, Rishikesh Ranade, Jay Pathak, and George~Em Karniadakis.
\newblock A hybrid iterative numerical transferable solver ({HINTS}) for pdes based on deep operator network and relaxation methods.
\newblock {\em arXiv preprint arXiv:2208.13273}, 2022.

\bibitem{chakraborty2025binned}
Dibyajyoti Chakraborty, Arvind~T Mohan, and Romit Maulik.
\newblock Binned spectral power loss for improved prediction of chaotic systems.
\newblock {\em arXiv preprint arXiv:2502.00472}, 2025.

\bibitem{goodfellow2014generative}
Ian Goodfellow, Jean Pouget-Abadie, Mehdi Mirza, Bing Xu, David Warde-Farley, Sherjil Ozair, Aaron Courville, and Yoshua Bengio.
\newblock Generative adversarial nets.
\newblock {\em Advances in neural information processing systems}, 27, 2014.

\bibitem{rezende2015variational}
Danilo Rezende and Shakir Mohamed.
\newblock Variational inference with normalizing flows.
\newblock In {\em International conference on machine learning}, pages 1530--1538. PMLR, 2015.

\bibitem{sohl2015deep}
Jascha Sohl-Dickstein, Eric Weiss, Niru Maheswaranathan, and Surya Ganguli.
\newblock Deep unsupervised learning using nonequilibrium thermodynamics.
\newblock In {\em International conference on machine learning}, pages 2256--2265. PMLR, 2015.

\bibitem{ho2020denoising}
Jonathan Ho, Ajay Jain, and Pieter Abbeel.
\newblock Denoising diffusion probabilistic models.
\newblock {\em Advances in neural information processing systems}, 33:6840--6851, 2020.

\bibitem{song2020score}
Yang Song, Jascha Sohl-Dickstein, Diederik~P Kingma, Abhishek Kumar, Stefano Ermon, and Ben Poole.
\newblock Score-based generative modeling through stochastic differential equations.
\newblock {\em arXiv preprint arXiv:2011.13456}, 2020.

\bibitem{karras2022elucidating}
Tero Karras, Miika Aittala, Timo Aila, and Samuli Laine.
\newblock Elucidating the design space of diffusion-based generative models.
\newblock {\em Advances in neural information processing systems}, 35:26565--26577, 2022.

\bibitem{saharia2022palette}
Chitwan Saharia, William Chan, Huiwen Chang, Chris Lee, Jonathan Ho, Tim Salimans, David Fleet, and Mohammad Norouzi.
\newblock Palette: Image-to-image diffusion models.
\newblock In {\em ACM SIGGRAPH 2022 conference proceedings}, pages 1--10, 2022.

\bibitem{robertson2023local}
Andreas~E Robertson, Conlain Kelly, Michael Buzzy, and Surya~R Kalidindi.
\newblock Local--global decompositions for conditional microstructure generation.
\newblock {\em Acta Materialia}, 253:118966, 2023.

\bibitem{robertson2024micro2d}
Andreas~E Robertson, Adam~P Generale, Conlain Kelly, Michael~O Buzzy, and Surya~R Kalidindi.
\newblock Micro2d: A large, statistically diverse, heterogeneous microstructure dataset.
\newblock {\em Integrating Materials and Manufacturing Innovation}, 13(1):120--154, 2024.

\bibitem{buzzy2024statistically}
Michael~O Buzzy, Andreas~E Robertson, and Surya~R Kalidindi.
\newblock Statistically conditioned polycrystal generation using denoising diffusion models.
\newblock {\em Acta Materialia}, 267:119746, 2024.

\bibitem{dong2024data}
Xinghao Dong, Chuanqi Chen, and Jin-Long Wu.
\newblock Data-driven stochastic closure modeling via conditional diffusion model and neural operator.
\newblock {\em arXiv preprint arXiv:2408.02965}, 2024.

\bibitem{huang2024diffusionpde}
Jiahe Huang, Guandao Yang, Zichen Wang, and Jeong~Joon Park.
\newblock Diffusion{PDE}: {G}enerative {PDE}-solving under partial observation.
\newblock {\em arXiv preprint arXiv:2406.17763}, 2024.

\bibitem{kohl2024benchmarking}
Georg Kohl, Liwei Chen, and Nils Thuerey.
\newblock Benchmarking autoregressive conditional diffusion models for turbulent flow simulation.
\newblock In {\em ICML 2024 AI for Science Workshop}, 2024.

\bibitem{wang2022deep}
Zhibo Wang, Xiangru Li, Luhan Liu, Xuecheng Wu, Pengfei Hao, Xiwen Zhang, and Feng He.
\newblock Deep-learning-based super-resolution reconstruction of high-speed imaging in fluids.
\newblock {\em Physics of Fluids}, 34(3), 2022.

\bibitem{molinaro2024generative}
Roberto Molinaro, Samuel Lanthaler, Bogdan Raoni{\'c}, Tobias Rohner, Victor Armegioiu, Zhong~Yi Wan, Fei Sha, Siddhartha Mishra, and Leonardo Zepeda-N{\'u}{\~n}ez.
\newblock Generative ai for fast and accurate statistical computation of fluids.
\newblock {\em arXiv preprint arXiv:2409.18359}, 2024.

\bibitem{lockwood2024generative}
Joseph~W Lockwood, Avantika Gori, and Pierre Gentine.
\newblock A generative super-resolution model for enhancing tropical cyclone wind field intensity and resolution.
\newblock {\em Journal of Geophysical Research: Machine Learning and Computation}, 1(4):e2024JH000375, 2024.

\bibitem{chandler2013invariant}
Gary~J Chandler and Rich~R Kerswell.
\newblock Invariant recurrent solutions embedded in a turbulent two-dimensional kolmogorov flow.
\newblock {\em Journal of Fluid Mechanics}, 722:554--595, 2013.

\bibitem{croitoru2023diffusion}
Florinel-Alin Croitoru, Vlad Hondru, Radu~Tudor Ionescu, and Mubarak Shah.
\newblock Diffusion models in vision: A survey.
\newblock {\em IEEE Transactions on Pattern Analysis and Machine Intelligence}, 45(9):10850--10869, 2023.

\bibitem{yang2023diffusion}
Ling Yang, Zhilong Zhang, Yang Song, Shenda Hong, Runsheng Xu, Yue Zhao, Wentao Zhang, Bin Cui, and Ming-Hsuan Yang.
\newblock Diffusion models: A comprehensive survey of methods and applications.
\newblock {\em ACM Computing Surveys}, 56(4):1--39, 2023.

\bibitem{thuerey2021physics}
Nils Thuerey, Philipp Holl, Maximilian Mueller, Patrick Schnell, Felix Trost, and Kiwon Um.
\newblock Physics-based deep learning.
\newblock {\em arXiv preprint arXiv:2109.05237}, 2021.

\bibitem{gupta2022towards}
Jayesh~K Gupta and Johannes Brandstetter.
\newblock Towards multi-spatiotemporal-scale generalized pde modeling.
\newblock {\em arXiv preprint arXiv:2209.15616}, 2022.

\bibitem{holl2024phiflow}
Philipp Holl and Nils Thuerey.
\newblock ${\Phi}_{\text{flow}}$ ({PhiFlow}): Differentiable simulations for {P}y{T}orch, {T}ensor{F}low and {J}ax.
\newblock In {\em International Conference on Machine Learning}. PMLR, 2024.

\bibitem{yeh2019resolvent}
Chi-An Yeh and Kunihiko Taira.
\newblock Resolvent-analysis-based design of airfoil separation control.
\newblock {\em Journal of Fluid Mechanics}, 867:572--610, 2019.

\bibitem{towne2023database}
Aaron Towne, Scott~TM Dawson, Guillaume~A Br{\`e}s, Adri{\'a}n Lozano-Dur{\'a}n, Theresa Saxton-Fox, Aadhy Parthasarathy, Anya~R Jones, Hulya Biler, Chi-An Yeh, Het~D Patel, et~al.
\newblock A database for reduced-complexity modeling of fluid flows.
\newblock {\em AIAA journal}, 61(7):2867--2892, 2023.

\bibitem{bres2017unstructured}
Guillaume~A Bres, Frank~E Ham, Joseph~W Nichols, and Sanjiva~K Lele.
\newblock Unstructured large-eddy simulations of supersonic jets.
\newblock {\em AIAA journal}, 55(4):1164--1184, 2017.

\bibitem{ovadia2023real}
Oded Ovadia, Vivek Oommen, Adar Kahana, Ahmad Peyvan, Eli Turkel, and George~Em Karniadakis.
\newblock Real-time inference and extrapolation via a {D}iffusion-inspired {T}emporal {T}ransformer {O}perator (ditto).
\newblock {\em arXiv preprint arXiv:2307.09072}, 2023.

\bibitem{settles2022schlieren}
Gary~S Settles and Alex Liberzon.
\newblock Schlieren and bos velocimetry of a round turbulent helium jet in air.
\newblock {\em Optics and Lasers in Engineering}, 156:107104, 2022.

\bibitem{wang2020denoising}
Phil Wang.
\newblock Denoising diffusion pytorch.
\newblock {\em GitHub repository}, 2020.

\bibitem{rahaman2019spectral}
Nasim Rahaman, Aristide Baratin, Devansh Arpit, Felix Draxler, Min Lin, Fred Hamprecht, Yoshua Bengio, and Aaron Courville.
\newblock On the spectral bias of neural networks.
\newblock In {\em International conference on machine learning}, pages 5301--5310. PMLR, 2019.

\bibitem{cao2019towards}
Yuan Cao, Zhiying Fang, Yue Wu, Ding-Xuan Zhou, and Quanquan Gu.
\newblock Towards understanding the spectral bias of deep learning.
\newblock {\em arXiv preprint arXiv:1912.01198}, 2019.

\bibitem{bowman2023spectral}
Benjamin Bowman.
\newblock {\em On the Spectral Bias of Neural Networks in the Neural Tangent Kernel Regime}.
\newblock University of California, Los Angeles, 2023.

\bibitem{basri2020frequency}
Ronen Basri, Meirav Galun, Amnon Geifman, David Jacobs, Yoni Kasten, and Shira Kritchman.
\newblock Frequency bias in neural networks for input of non-uniform density.
\newblock In {\em International conference on machine learning}, pages 685--694. PMLR, 2020.

\bibitem{stein1971introduction}
Elias~M Stein and Guido Weiss.
\newblock {\em Introduction to Fourier analysis on Euclidean spaces}, volume~1.
\newblock Princeton university press, 1971.

\bibitem{song2019generative}
Yang Song and Stefano Ermon.
\newblock Generative modeling by estimating gradients of the data distribution.
\newblock {\em Advances in neural information processing systems}, 32, 2019.

\bibitem{proakis2007digital}
John~G Proakis.
\newblock Digital signal processing: principles, algorithms, and applications, 4/e.
\newblock {\em Pearson Education India}, 2007.

\end{thebibliography}
\newpage

\appendix
\counterwithin{figure}{section}
\setcounter{figure}{0}

\section{POD Analysis}
\label{app: pod_analysis}

\begin{figure}[H]
  \centering
  \includegraphics[width=0.6\textwidth]{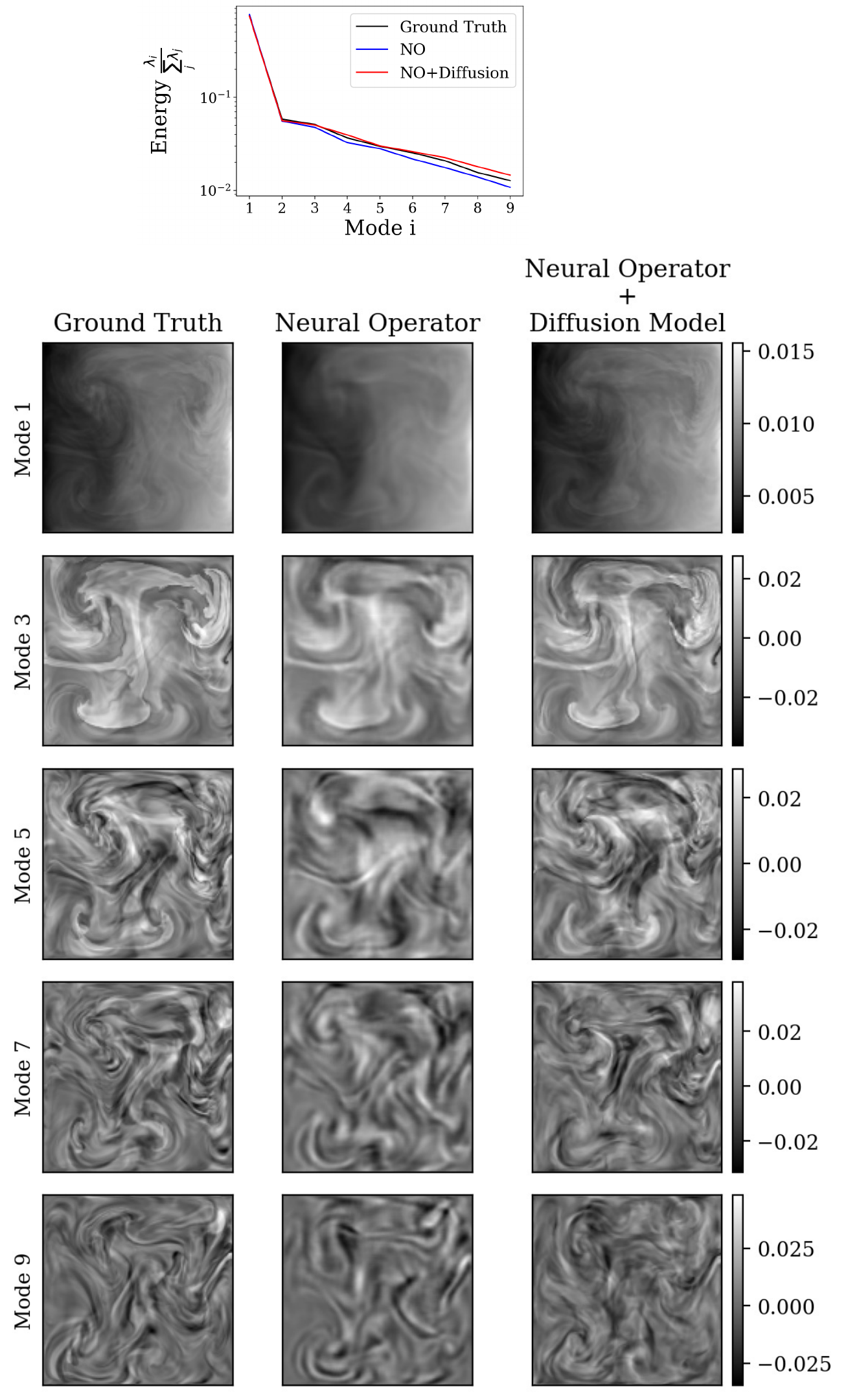}
  \caption{\textbf{POD analysis of buoyancy-driven flow.} We compare the decay of eigenvalues obtained from performing proper orthogonal decomposition of 10 snapshots predicted by the neural operator (UNet) and diffusion model conditioned on the neural operator with the ground truth solution. 
  Additionally, we also visualize and compare POD modes corresponding to $i=1,3,5,7,9$. }
  \label{fig: pod_buoyancy}
\end{figure}

\begin{figure}[H]
  \centering
  \includegraphics[width=0.99\textwidth]{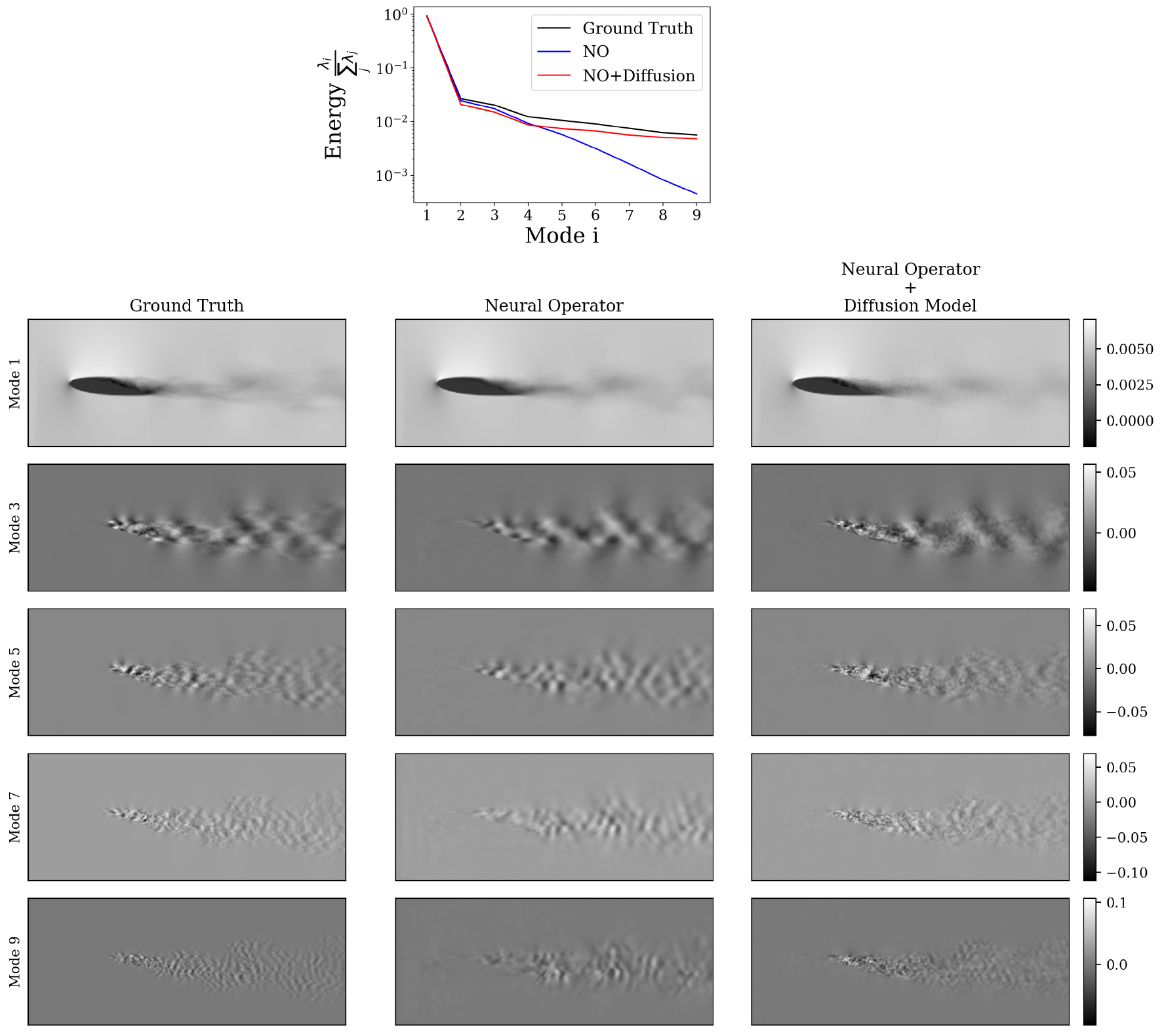}
  \caption{\textbf{POD analysis of a turbulent airfoil.}
  We compare the decay of eigenvalues obtained from performing proper orthogonal decomposition of 10 snapshots predicted by the neural operator (TC-UNet) and diffusion model conditioned on the neural operator with the ground truth solution. 
  Additionally, we also visualize and compare POD modes corresponding to $i=1,3,5,7,9$. }
  \label{fig: pod_airfoil}
\end{figure}

In the main manuscript, we investigate the energy spectra with respect to the wavenumber corresponding to the neural operator and diffusion model conditioned on the neural operator, computed on each time step separately.
Here we perform a detailed proper orthogonal decomposition (POD) analysis by performing singular value decomposition on 10 consecutive timesteps of the trajectory, and plot the eigen energy spectrum.
Additionally, we also visualize the corresponding eigen modes/basis required for accurate reconstruction of the aforementioned timesteps.
The POD analysis for Kolmogorov flow \autoref{fig: pod_kolmogorov}, buoyancy driven transport \autoref{fig: pod_buoyancy}, turbulent airfoil \autoref{fig: pod_airfoil}, 3D turbulent jet \autoref{fig: pod_jet_3d} and schlieren velocimetry \autoref{fig: pod_schlieren} are shown below.


\newpage

\newpage

\newpage
\begin{figure}[H]
  \centering
  \includegraphics[width=0.99\textwidth]{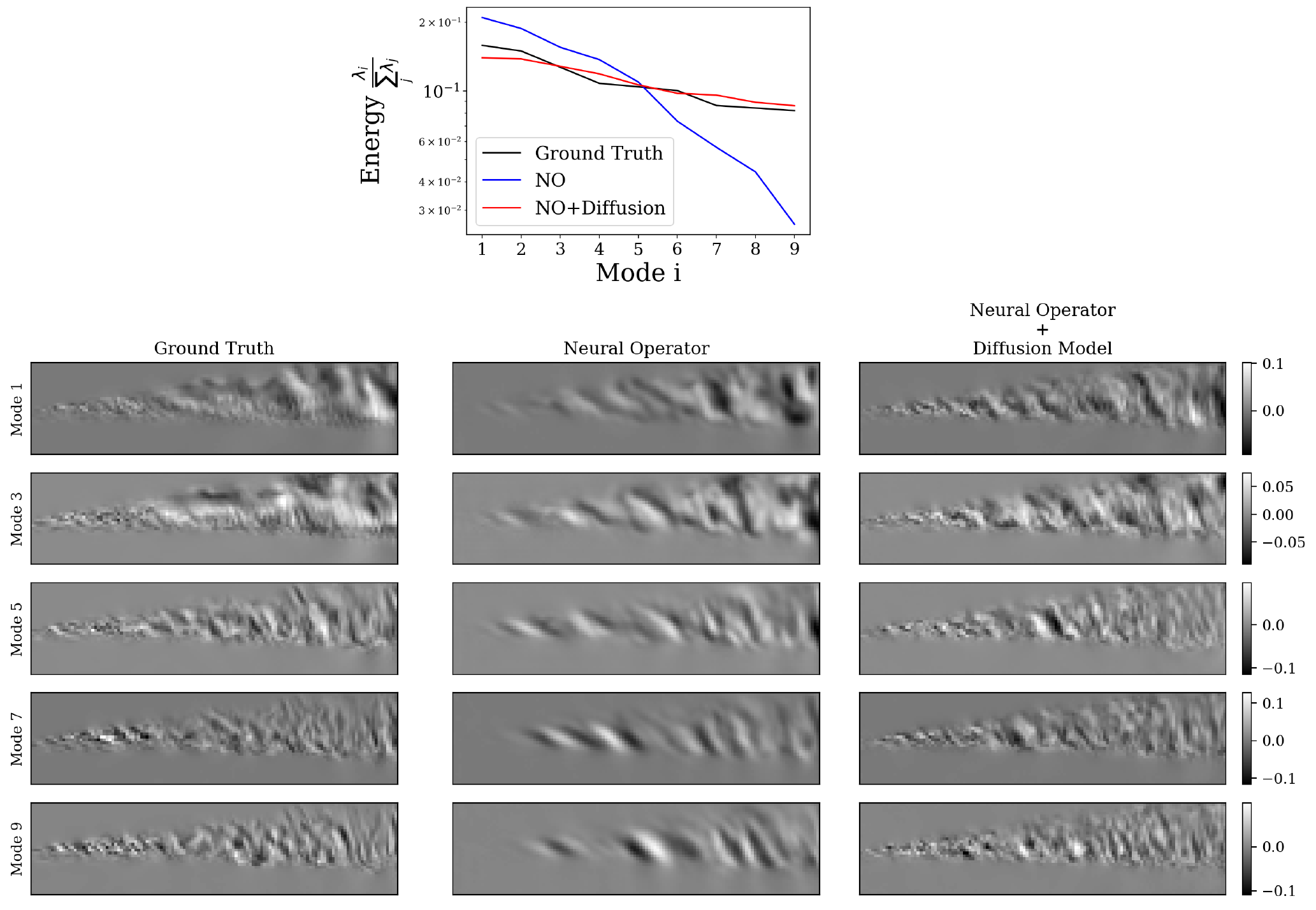}
  \caption{\textbf{POD analysis of a 3D jet}
  We perform POD on the angular velocity component at $\theta=0$. 
  We compare the decay of eigenvalues obtained from performing proper orthogonal decomposition of 10 snapshots predicted by the neural operator (TC-UNet) and diffusion model conditioned on the neural operator with the ground truth solution. 
  Additionally, we also visualize and compare POD modes corresponding to $i=1,3,5,7,9$. }
  \label{fig: pod_jet_3d}
\end{figure}

\newpage

\begin{figure}[H]
  \centering
  \includegraphics[width=0.99\textwidth]{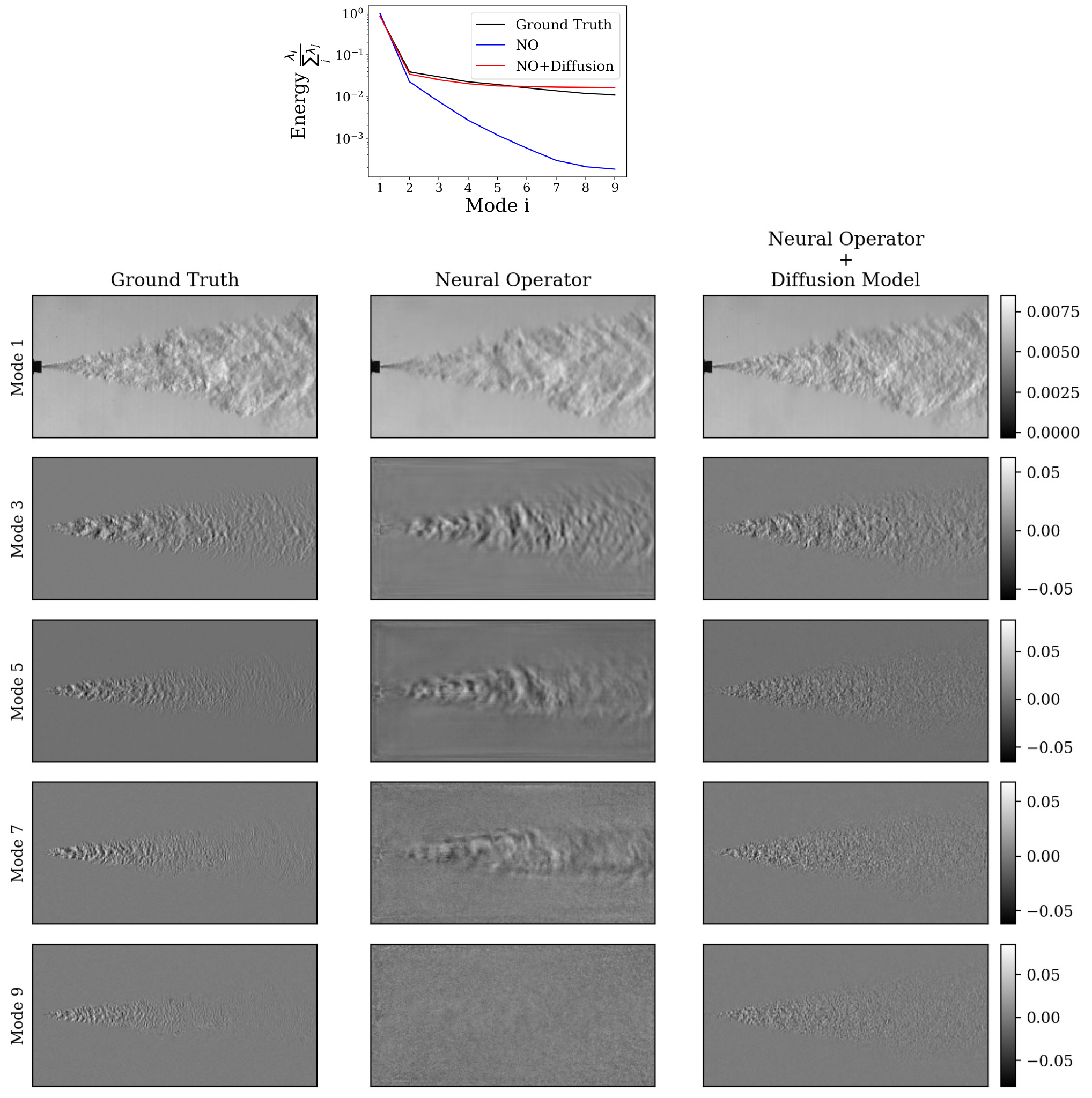}
  \caption{\textbf{POD analysis of Schlieren velocimetry.} We compare the decay of eigenvalues obtained from performing proper orthogonal decomposition of 10 snapshots predicted by the neural operator and diffusion model conditioned on neural operator (TC-UNet) with the true solution obtained from Schlieren velocimetry. 
  Additionally, we also visualize and compare POD modes corresponding to $i=1,3,5,7,9$. }
  \label{fig: pod_schlieren}
\end{figure}

\end{document}